\documentclass{article}

\usepackage{microtype}
\usepackage{graphicx}
\usepackage{subcaption}
\usepackage{booktabs} % for professional tables

\usepackage{hyperref}

\usepackage[accepted]{icml2026}

\usepackage{amsmath}
\usepackage{bm}
\usepackage{amssymb}
\usepackage{mathtools}
\usepackage{amsthm}
\usepackage{tabularx}
\usepackage{array}
\usepackage[table]{xcolor}

\usepackage{algorithm}
\usepackage{algpseudocode}
\usepackage{amsmath,amssymb}

\newcolumntype{H}{>{\setbox0=\hbox\bgroup}c<{\egroup}@{}}

\newcommand{\piold}{\pi_{\text{old}}}

\newcommand{\pitheta}{\pi_{\theta}}

\newcommand{\M}{\text{M}}

\newcommand{\ours}[0]{LaViDa-R1}

\usepackage[capitalize,noabbrev]{cleveref}

\theoremstyle{plain}

\theoremstyle{definition}

\theoremstyle{remark}

\usepackage[textsize=tiny]{todonotes}

\icmltitlerunning{\ours}

\begin{document}

\twocolumn[
  \icmltitle{\ours: Advancing Reasoning for Unified Multimodal Diffusion Language Models}

  % It is OKAY to include author information, even for blind submissions: the
  % style file will automatically remove it for you unless you've provided
  % the [accepted] option to the icml2026 package.

  % List of affiliations: The first argument should be a (short) identifier you
  % will use later to specify author affiliations Academic affiliations
  % should list Department, University, City, Region, Country Industry
  % affiliations should list Company, City, Region, Country

  % You can specify symbols, otherwise they are numbered in order. Ideally, you
  % should not use this facility. Affiliations will be numbered in order of
  % appearance and this is the preferred way.
  \icmlsetsymbol{equal}{*}

  % \begin{icmlauthorlist}
  %   \icmlauthor{Firstname1 Lastname1}{equal,yyy}
  %   \icmlauthor{Firstname2 Lastname2}{equal,yyy,comp}
  %   \icmlauthor{Firstname3 Lastname3}{comp}
  %   \icmlauthor{Firstname4 Lastname4}{sch}
  %   \icmlauthor{Firstname5 Lastname5}{yyy}
  %   \icmlauthor{Firstname6 Lastname6}{sch,yyy,comp}
  %   \icmlauthor{Firstname7 Lastname7}{comp}
  %   %\icmlauthor{}{sch}
  %   \icmlauthor{Firstname8 Lastname8}{sch}
  %   \icmlauthor{Firstname8 Lastname8}{yyy,comp}
  %   %\icmlauthor{}{sch}
  %   %\icmlauthor{}{sch}
  % \end{icmlauthorlist}
\begin{icmlauthorlist}
  Shufan Li$^{1,2,*, \dagger}$, Yuchen Zhu$^{1,3,*, \dagger}$, Kangning Liu$^{1}$, Zhe Lin$^{1}$ \\ Yongxin Chen$^{3}$, Molei Tao$^{3}$, Aditya Grover$^{2}$,  Jiuxiang Gu$^{1}$, Jason Kuen$^{1}$ \\
$^1$Adobe~~$^2$UCLA~$^3$Georgia Tech \\
* Equal Contribution  \; $\dagger$ Work done primarily during internship at Adobe Research\\
~~ \\
~~ \\

\end{icmlauthorlist}

  % \icmlaffiliation{yyy}{Department of XXX, University of YYY, Location, Country}
  % \icmlaffiliation{comp}{Company Name, Location, Country}
  % \icmlaffiliation{sch}{School of ZZZ, Institute of WWW, Location, Country}

  \icmlcorrespondingauthor{Shufan Li}{jacklishufan@cs.ucla.edu}
  \icmlcorrespondingauthor{Yuchen Zhu}{yzhu738@gatech.edu}
  \icmlcorrespondingauthor{Jiuxiang Gu}{jigu@adobe.com}
  \icmlcorrespondingauthor{Jason Kuen}{kuen@adobe.com}

  % You may provide any keywords that you find helpful for describing your
  % paper; these are used to populate the "keywords" metadata in the PDF but
  % will not be shown in the document
  \icmlkeywords{Machine Learning, ICML}
\centering
  \vbox{
\includegraphics[width= 0.85\textwidth]{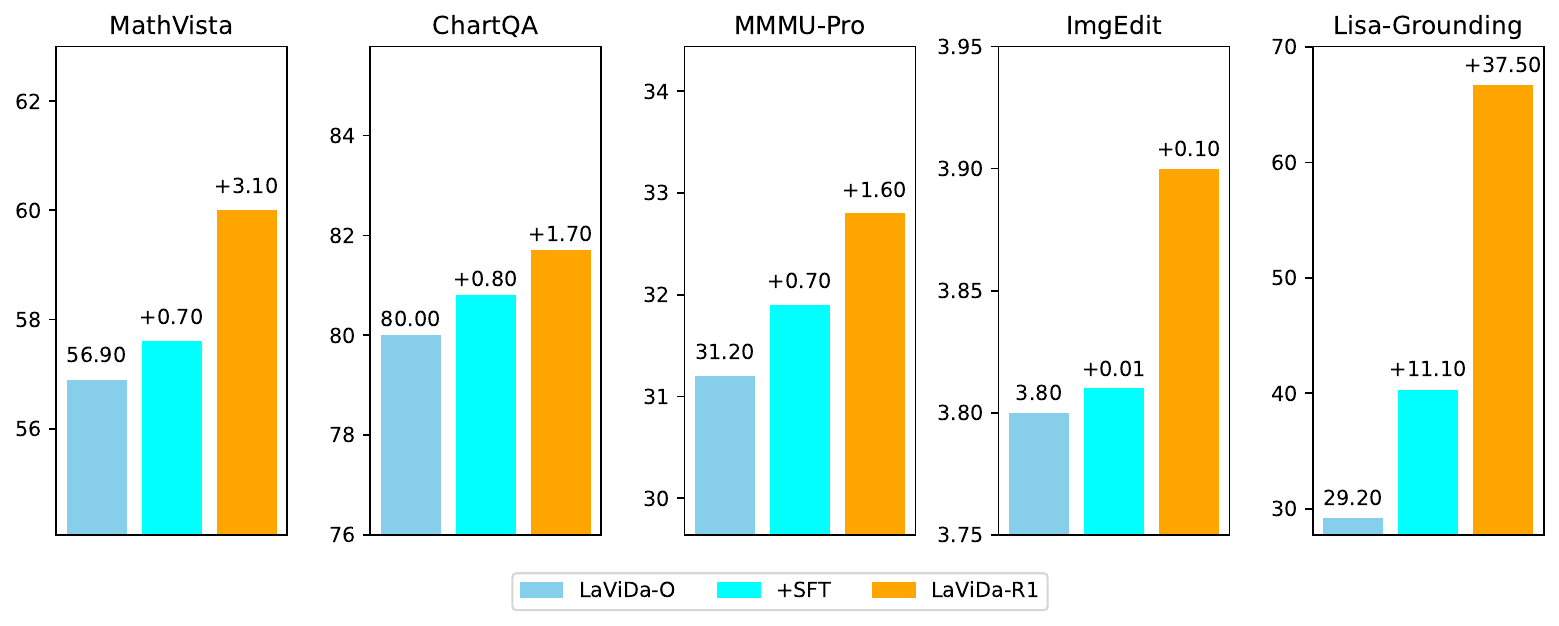}
\captionof{figure}{\textbf{We introduce \ours}, a multimodal diffusion language model with strong reasoning capabilities across diverse tasks. \ours incorporates a novel unified post-training that significantly improves upon the base model LaViDa-O \cite{li2025lavidao} and SFT baseline on visual math reasoning, visual question answering, image editing, and object grounding tasks. }
\label{fig:teaser} 
\vspace{-1em}
}

  \vskip 0.3in

]

% this must go after the closing bracket ] following \twocolumn[ ...

% This command actually creates the footnote in the first column listing the
% affiliations and the copyright notice. The command takes one argument, which
% is text to display at the start of the footnote. The \icmlEqualContribution
% command is standard text for equal contribution. Remove it (just {}) if you
% do not need this facility.

% Use ONE of the following lines. DO NOT remove the command.
% If you have no special notice, KEEP empty braces:
\printAffiliationsAndNotice{}  % no special notice (required even if empty)
% Or, if applicable, use the standard equal contribution text:
% \printAffiliationsAndNotice{\icmlEqualContribution}

\begin{abstract}
Diffusion language models (dLLMs) recently emerged as a promising alternative to auto-regressive LLMs. The latest works further extended it to multimodal understanding and generation tasks. In this work, we propose \ours, a multimodal, general-purpose reasoning dLLM. Unlike existing works that build reasoning dLLMs through task-specific reinforcement learning, \ours~incorporates diverse multimodal understanding and generation tasks in a unified manner. In particular, \ours~is built with a novel unified post-training framework that seamlessly integrates supervised finetuning (SFT) and multi-task reinforcement learning (RL). It employs several novel training techniques, including answer-forcing, tree search, and complementary likelihood estimation, to enhance effectiveness and scalability. Extensive experiments demonstrate \ours's strong performance on a wide range of multimodal tasks, including visual math reasoning, reason-intensive grounding, and image editing.
 
\end{abstract}

% \shufan{Figures: 1. Teaser, Bar Chart, Figure 2 , Explainer, pipeline, Figure 3,  Qualitative Results,Figure 4, Some Reward Curve  }
\section{Introduction}

Unified Multimodal Large Language Models (MLLMs) such as GPT-4o \cite{openai2024gpt4o} have demonstrated strong utility on diverse scenarios. Traditionally, these models are built as auto-regressive (AR) models that generate tokens sequentially. Recently, dLLMs have emerged as a promising alternative to auto-regressive models in many language \cite{nie2025large,dream2025} and multimodal tasks \cite{yang2025mmada,li2025lavidao,li2025lavida,swerdlow2025unidisc}. Instead of generating tokens in a left-to-right order, dLLMs start with a fully masked sequence and gradually unmask it through multiple diffusion steps, decoding multiple tokens in parallel. Compared with AR models, dLLMs offer many attractive properties such as faster inference speed \cite{wu2025fast}, bi-directional context \cite{li2025lavida,nie2025large}, and a unified generation paradigm for visual and text tokens \cite{li2025lavidao,yang2025mmada}.

To improve the performance of AR models, a common technique is to incorporate a reasoning process \cite{guo2025deepseek}, in which the model generates text-reasoning traces before producing the final output. This approach has shown to be highly effective on complex tasks such as  math reasoning \cite{shao2024deepseekmath} and coding \cite{li2025structured}. Recent work \cite{shen2025vlm,deng2025emerging} further extends the reasoning process to support multimodal understanding and generation tasks. 

The reasoning capability of a model is typically acquired via post-supervised finetuning on Chain-of-Thought (CoT) data \cite{wei2022chain}  and reinforcement learning (RL) \cite{guo2025deepseek,yang2025qwen3}. While these approaches were first developed for AR models, recent work has explored applying them to build reasoning dLLMs and multimodal dLLMs \cite{yang2025mmada,zhao2025d1,zhu2025enhancing}. While AR models suffer from linear error accumulation due to causal masking, reasoning dLLMs leverage global visibility. By jointly modeling reasoning and results, they enable holistic refinement, allowing the emerging answer to adjust the reasoning trace and providing a unified framework effective for both spatial visual modeling and complex logical reasoning. 

%Compared with AR counterparts, reasoning dLLMs have demonstrated several advantages, such as bi-directional attention and faster decoding \cite{wu2025fast, hong2025wide}.

While the reasoning dLLM literature has seen some progress, several key challenges remain unaddressed. First, existing work focuses on a limited set of tasks, such as mathematical reasoning, and often requires dataset-specific fine-tuning. Extending RL to build a general-purpose reasoning model that supports diverse, multimodal tasks such as image editing and reason-intensive object grounding remains largely unexplored. Second, training dLLM with reinforcement learning is prone to collapse even in the presence of KL divergence regularizer. Furthermore, incorporating the KL-regularizer hinders models' exploration during training, thereby degrading task performance. Third, for complex or difficult tasks, the model may fail to generate high-quality samples during training, leading to a low-quality training signal or, in the worst case, a zero training signal due to diminishing returns. Finally, unlike AR models, which can evaluate sequence likelihood exactly and efficiently, computing sequence likelihood for dLLMs is intractable and is typically estimated via the Monte Carlo (MC) method. This approach poses unique challenges for training stability as it produces missing and imbalanced token gradients.

To address these gaps, we propose LaViDa-R1, a recipe for building strong-performing multimodal dLLMs. Compared with existing methods, LaViDa-R1 introduces several key innovations. First, it introduces a unified framework that encompasses a diverse range of visual and language tasks, including mathematical reasoning, visual question answering, reason-intensive grounding, and image editing. Second, it introduces a novel post-training objective that seamlessly integrates SFT and the RL paradigm. By replacing the KL divergence term with SFT regularization, LaViDa-R1 allows the model to sufficiently explore beyond the distribution of a pretrained base model while also preventing collapse. Third, to address the lack of a training signal when no high-quality samples are generated for difficult prompts, we incorporate two guided rollout generation mechanisms to construct high-quality samples. When ground-truth answers are available, we employ an answer-forcing mechanism that leverages dLLMs' inpainting capabilities to artificially construct high-quality reasoning traces on the fly. When the ground-truth answer is unavailable, we employ a tree-search algorithm that tailors the generated distribution towards higher-quality outputs. Finally, we propose a complementary likelihood estimator that improves upon existing MC methods by addressing the missing-signal and imbalanced-gradient problems discussed above.  

To validate the effectiveness of \ours, we conducted extensive experiments covering a wide range of tasks. Results show that \ours~achieves strong reasoning performances on multiple benchmarks such as MathVerse, ChartQA, Lisa-Grounding, and ImgEdit. 
\section{Background and Related Works}

\subsection{Discrete diffusion models}

Early works on discrete diffusion models ~\citep{austin2021structured-d3pm,sahoo2024simple,lou2023discrete-sedd, ou2024your, shi2024simplified} first developed principled frameworks for training and sampling from masked generative models (MGMs) by formalizing the unmasking process of MGMs as a discrete diffusion process. Later works, such as Mercury and LLaDA~\cite{khanna2025mercury,nie2025large,dream2025}, scaled discrete diffusion models to large-scale language modeling, achieving performance comparable to autoregressive LLMs while offering benefits such as bidirectional context and faster inference.  Recent works such as LaViDa-O and MMaDa~\citep{li2025lavida,li2025lavidao,yu2025dimple,shi2025muddit,yang2025mmada} further expanded dLLMs to multimodal understanding and generation tasks.

% TODO, fix for pi, x, y format 
% where its $i$-th token is denoted as $\bm{y}_0[i]$
Formally, given a sequence $\bm{y}_0$ of length $L$, and a conditional prompt $\bm{x}$, the forward masked diffusion process $q(\bm{y}_t|\bm{y}_s)$ progressively mask tokens over the time interval $[0,1]$, with $1 \ge t \ge s \ge 0$. At $t = 0$,  no tokens are masked. At $t = 1$, the sequence $\bm{y}_1 = [\M, \M, \cdots, \M]$ consists entirely of a special mask token $\M$. When $0 < t < 1$, $\bm{y}_t$ contains a mixture of clean and masked tokens. A dLLM policy model $\pi_\theta$ is trained to model the reverse process $p(\bm{y}_s|\bm{y}_t,\bm{x})$. The masked diffusion objective is defined as:
\begin{align}
    \mathcal{L}_{\text{dLLM}}(\theta) = -\mathbb{E}_{t, \bm{y}_0, \bm{y}_t,
    \bm{x}} \left[\frac{1}{t} \log \pi_\theta(\bm{y}_0 | \bm{y}_t,\bm{x})\right]
\end{align}
% where $\pi_\theta(\bm{y}_0 | \bm{y}_t,\bm{x})$ is factorized as $\prod_{i=1}^L \pi_\theta(\bm{y}_0[i] | \bm{y}_t,\bm{x})$ following standard independence assumptions~\citep{sahoo2024simple}.  
where $\pi_\theta(\bm{y_0} | \bm{y_t},\bm{x})$ is factorized to the product of per-token distribution  $\prod_{i=1}^L \pi_\theta(\bm{y_0}[i] | \bm{y_t},\bm{x})$  \citep{sahoo2024simple}. At inference, given a prompt $\bm{x}$, we initialize with a fully masked sequence $\bm{y}_1$ and iteratively apply the learned reverse process  $\pitheta$ to progressively unmask tokens until a clean sequence $\bm{y}_0$ is obtained.

% Most existing MDMs adopt a dense parameterization. At intermediate steps where $0 < t < 1$, all $L$ tokens in $X_t$, both masked and unmasked, are passed to the neural network $p_\theta$, which outputs a dense tensor $y \in \mathbb{R}^{L \times V}$, where $V$ is the vocabulary size. Each $y[i] \in \mathbb{R}^{V}$ represents the logits corresponding to $\log p_\theta(X_0^i | X_t)$.  
% This design is computationally inefficient because all $L$ tokens must be processed even when predictions are only needed for a small subset.  

\subsection{Reinforcement Learning}
Reinforcement learning  \cite{schulman2017proximal} can effectively improve the reasoning capability of LLMs. GRPO \cite{shao2024deepseekmath} is one of the best-performing RL methods,  whose objective has the following form
\begin{align}
\label{eq:grpo}
& J_{\text{grpo}}(\theta) = {\mathbb{E}}\Big[\frac{1}{N} \sum_{i = 1}^{N} \min\Big(\dfrac{\pitheta(\bm{y}^i|\bm{x})}{\piold(\bm{y}^i|\bm{x})}A_i^{\text{GRPO}
}, \notag \\
&\quad \operatorname{clip}\big(\dfrac{\pitheta(\bm{y}^i|\bm{x})}{\piold(\bm{y}^i|\bm{x})}, 1-\varepsilon, 1 + \varepsilon\big) A^{\text{GRPO}}_i\Big) 
-\beta\operatorname{kl}(\bm{y}^i)\Big], \notag\\
& \qquad A^{\text{GRPO}}_i = \frac{r_i - \operatorname{mean}(r_1, \dots, r_N)}{\operatorname{std}(r_1, \dots, r_N)}
\end{align}
where $\bm{y}^1, \dots, \bm{y}^N$ is a group of $N$ responses sampled with prompt $\bm{x}$, $\operatorname{kl}(\bm{y}^i)$ is a per-sample reverse KL estimator, $r_i$ is the per-sample reward and $A^{\text{GRPO}}_i$ is the per-sample advantage, which is the normalized reward.

It has been shown that in a pure on-policy setting, GRPO advantages can be simplified as
\begin{align}
J(\theta) = \mathbb{E}\Big[\frac{1}{N} \sum_{i = 1}^{N} A_{i}^{\text{GRPO}} \log \pi_\theta(\bm{y}^i | \bm{x}) - \beta \operatorname{kl}(\bm{y}^i)\Big]
% & \log \pitheta(\bm{y} |\bm{x}) = \mathbb{E}_{t, \bm{y}_t}\Big[-w(t)\sum_{k: \bm{y}_t[k]=\M} \log \pitheta(\bm{y}[k] |\bm{y}_t, \bm{x})\Big]
\label{eq:grpo_online}
\end{align}

Interestingly, it is shown that by changing $\beta$, and the definition of $A_{i}$ \cite{shao2024deepseekmath}, we can use the same form to represent many other objectives such as SFT and DPO \cite{rafailov2023direct}. Inspired by this view, our work proposes a practical unified post-training method that combines SFT, RL, and self-distillation loss to improve reasoning capabilities.

\textbf{RL for dLLMs. }Multiple works have also explored applying GRPO-style RL to dLLMs \cite{zhao2025d1,gong2025diffucoder,wang2025d2,tang2025wd1,zhu2025enhancing,wang2025revolutionizing,ou2025principled,zheng2025group}, mostly focused on language-only tasks with task-specific training. Few works have explored applying RL to multimodal tasks. Uni-GRPO \cite{yang2025mmada} first explored extending RL to improve math reasoning, image captioning and text-to-image generation simultaneously. Our work extends RL to a broader set of tasks such as reason-intensive object grounding and image editing. We provide a more thorough review to these literature in Appendix \ref{sec:appendix_related}.

\textbf{RL for Multimodal tasks.} For AR VLMs and unified MLLMs, many works explored enhancing reasoning with RL \cite{wang2025vl, shen2025vlm, meng2025mm, zhou2025r1, yang2025r1, deng2025openvlthinker, huang2025vision, wang2025sota, yuan2025vl}, achieving successes on a wide range of visual understanding and generation tasks.  We also note that while RL is commonly associated with reasoning, it can also be applied to improve visual generation tasks without reasoning elements. Multiple works explored applying RL in image-output-only setup (e.g. Stable Diffusion) to improve text-to-image generation and image-editing quality \cite{li2025uniworld, geng2025x, wei2025skywork, liu2025flow, zheng2025diffusionnft, wu2025qwen, luo2025editscore}. Most existing work focuses on applying RL techniques in a task-specific manner, with a few exceptions exploring unified reasoning for unified multimodal models \cite{tian2025unigen, xin2025lumina, yang2025mmada, cui2025emu3}. Our work focuses specifically on improving multimodal task performance by enhancing \textit{reasoning} capabilities using RL. It is more closely aligned with the VLM and MLLM reasoning literature than with the general RL literature on visual generation.

\subsection{Improve reasoning with non-online-RL methods}
Other lines of work address the challenge of LLM alignment beyond GRPO-style policy-gradient methods. Direct Preference Optimization (DPO) \cite{rafailov2023direct, azar2024general, zhao2023slic} aligns LLMs using off-line paired preference data, and Online-DPO \cite{guo2024direct} optimizes models based on preference pairs obtained from model-generated responses and an external reward model as a judge. Self-play \cite{chen2024self, wu2024self, rosset2024direct, swamy2024minimaximalist} formulate the task of LLM alignment as a two-player game. Self-distillation-type methods \cite{yang2024faster, amini2024variational} progressively improve the model by distilling optimized self-generated rollouts back into the model.

Specifically, BOND \cite{sessa2024bond} iteratively distilled the Best-of-N model output distribution into the model policy by minimizing the KL divergence between them, which is equivalent to performing SFT on the best sequence $\bm{y}^j$ among a group of responses $\bm{y}^1, \dots, \bm{y}^N$ from the same prompt $\bm{x}$, with the following objective
\begin{align}
J_{\text{distill}}(\theta) = \log \pi_\theta(\bm{y}^j | \bm{x}), ~ j = \operatorname{argmax}_i ~ r_i
\label{eq:distill}
\end{align}
Our work creatively combines the best-of-N distillation objective with the standard RL objective. We also explored alternatives like Online-DPO. Further discussions can be found in Appendix \ref{sec:appendix_unified}.

\begin{figure*}[t]
\vspace{-1em}
    \centering
    \includegraphics[width=1.0\linewidth]{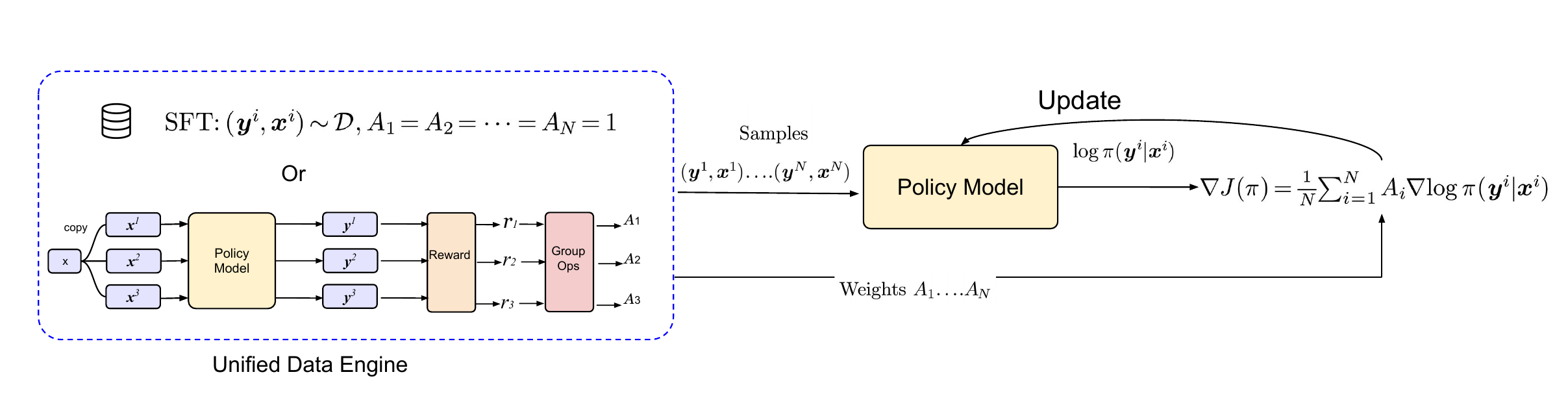}
    \caption{\textbf{Unified Post Training Framework of \ours.} At each training step, a generic data engine provides prompts-response pairs of $(\bm{y}^i,\bm{x}^i)$, and sample weights $A_i$, either by loading from a dataset or by online generation. The policy model is then used to compute the log-likelihood of each sequence $\log \pi_\theta(\bm{y}^i|\bm{x}_i)$. Finally, we optimize the proposed unified policy gradient objective. }
    \label{fig:rl_pipeline}
\end{figure*}

\section{Method}
 In this section, we introduce the training framework for LaViDa-R1, which comprises three main components: a unified post-training policy-gradient objective, guided rollout-generation algorithms that efficiently sample high-quality outputs during training, and a novel, stable, complementary-masking-based likelihood estimator. 

\subsection{Unified Post-training}
\label{sec:method_unified_training}
As is first noted in \cite{shao2024deepseekmath}, many post-training objectives, including online GRPO \cite{shao2024deepseekmath} (without KL regularization), Online DPO \cite{guo2024direct}, SFT, and self-distillation \cite{sessa2024bond}, can be written in the same form of policy gradient objectives: 
\begin{align}
\label{eq:awc_loss}
    J_\text{Unified}(\theta) = \frac{1}{N} \sum_{i = 1}^{N} A_i \log \pi_\theta(\bm{y}^i | \bm{x}^i)
\end{align}
The design choices that differentiate these objectives are the sources of $(\bm{y^i},\bm{x^i})$ pairs and the per-sample weights $A_i$. For example,

\begin{itemize}
    \item When $\bm{y}^i \sim \pitheta(\cdot | \bm{x}^i)$ is sampled form the policy model during training, $A_i = A_{i}^{\text{GRPO}}$, $J(\theta)$ is the online GRPO objective with zero KL regularization from Eq. \ref{eq:grpo_online}.
    \item  When $\bm{y}^i \sim \pitheta(\cdot | \bm{x}^i)$, $A_i = A_i^{\text{distill}}$, with $A_i^{\text{distill}} = 1$ if $i = \arg \max (r_1, \dots, r_N)$ and 0 otherwise, 
$J(\theta)$ is equivalent to the best-of-N distillation (Eq. \ref{eq:distill}).
    \item When $(\bm{y}^i,\bm{x}^i) \sim \mathcal{D}$ comes from a offline training dataset, $A_i = A_{i}^{\text{sft}} = 1$, $J(\theta)$ is the SFT loss.
\end{itemize}

% 
    % \item When $\bm{y}^i \sim \pitheta(\cdot | \bm{x}^i)$, $A_i = A_{i}^{\text{DPO}}$, where it's defined for a pair of index $i, j$ such that $r_i > r_j$, and the advantage is defined as,
    % \begin{align*}
    % \resizebox{0.95\columnwidth}{!}{$
    %     A^{\text{DPO}}_{i} = - A^{\text{DPO}}_j =  \sigma\Big(\beta \Big[\log \dfrac{\piref(\bm{y}^i |\bm{x^i})}{\piref(\bm{y}^j|\bm{x}^j)}   - \log \dfrac{\pitheta(\bm{y}^i |\bm{x^i})}{\pitheta(\bm{y}^j|\bm{x}^j)}\Big]\Big)$
    % }
    % \end{align*}
    % In such case, $J(\theta)$ is equivalent to the online DPO objective \cite{guo2024direct}. 
% Deepseek-Math\cite{shao2024deepseekmath} first showed that many post training techniques including SFT and  GRPO have the following gradient form

% \begin{equation}
%     \nabla J_\theta= \frac{1}{N}\sum_{i=1}^NA_i  \nabla \log \pi_\theta(y_i|x_i)
%     \label{eq:advantages}
% \end{equation}

% Where $A_i$ is a weight term depending of the specific algorithm. For on-policy GRPO, $A_i$ is just advantages $a_i$ from Eq\shufan{todo,callback to related}. For SFT, $A_i$ is just 1.  While this view was originally proposed to provide unified analysis of different post-training methods, it is not hard to see that This gives a surrogate loss by integrating both sides of the equation. 

% \begin{equation}
%     J_\theta= \frac{1}{N}\sum_{i=1}^NA_i  \log \pi_\theta(y_i|x_i)
%     \label{eq:awc_loss}
% \end{equation}

This observation has several important implications. First, we can easily combine training batches of these objectives by concatenating lists of $(\bm{y}^i,\bm{x}^i)$ and corresponding $A_i$.  This pipeline is concretely shown in Figure \ref{fig:rl_pipeline}. At each training step, a generic data engine provides pairs of prompts and responses $(\bm{y}^i,\bm{x}^i)$, as well as corresponding advantage values $A_i$. These can be obtained either by loading from an offline training dataset or by online generation, followed by reward and advantage calculation such as $A^{\text{distill}}_i$ or $A^{\text{GRPO}}_i$. The policy model is then used to compute the log-likelihood of each sequence $\log \pi_\theta(\bm{y}^i|\bm{x}_i)$. Finally, we optimize the unified objective in Eq. \ref{eq:awc_loss}. This design is illustrated in Fig \ref{fig:rl_pipeline}.

Second, for on-policy objectives where $\bm{y}^i \sim \pitheta(\cdot | \bm{x}^i)$ are sampled from the policy model, we can efficiently combine different objectives by simply aggregating the advantage values from each method using a weighted average, without the need to resample rollouts across different losses. For example, we can simultaneously perform online GRPO and best-of-N self-distillation by adopting a new advantage $A_i^{\text{aggr}} =\gamma A_i^\text{distill} + (1-\gamma) A_i^\text{GRPO}$ for each sample with barely any additional computational overhead.

In our final design, we combined SFT, online GRPO and online self-distillation objectives, with $\gamma=0.5$. We also explored other objectives that can be written in this form such as online DPO and SLiC \cite{rafailov2023direct,zhao2023slic}. Further details are provided in Appendix \ref{sec:appendix_unified}. 

Intuitively, adding the SFT objective can serve as a substitute for KL regularization. It allows the model to sufficiently explore the action space without being constrained by a suboptimal reference model, while preventing collapse. Furthermore, from a computational-efficiency perspective, removing the need for a reference model significantly reduces the cost of RL training, since we no longer need to load it into GPU memory or host it on a separate server. On the other hand, incorporating a self-distillation objective amplifies the training signal from the best sample in the group, leading to stronger training signals.

\subsection{Guided Rollout Generation}
Online RL with group-based advantage computation is known to suffer from a vanishing training signal when all generated responses receive low rewards, resulting in zero advantage for all responses and rendering the RL process ineffective. To effectively address this notorious issue, we propose using guided generation to create high-quality rollout samples during training. We consider two types of guided generation algorithms, each with different operating scenarios. Answer-forcing is applied when we have access to the ground-truth answers to the training questions (e.g. math reasoning). When answers are unavailable, we resort to tree search, which is applicable when a real-valued reward function is available.

\subsubsection{Answer Forcing}
\label{sec:method_answer_forcing}
\begin{figure}
    \centering
    \includegraphics[width=1.0\linewidth]{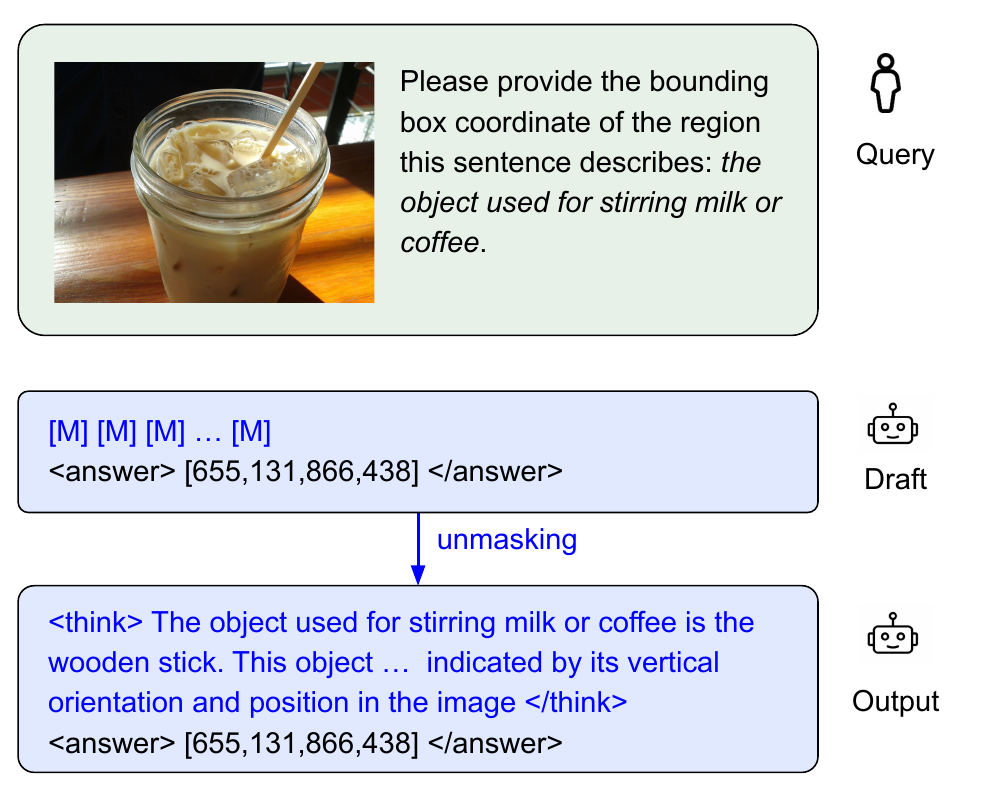}
    \caption{\textbf{Answer-Forcing.} We initialize a partially masked sequence with ground truth answer injected at the end, and use the diffusion unmasking process to obtain the reasoning trace.}
    \label{fig:answer-forcing}
    \vspace{-1em}
\end{figure}

We leverage dLLMs' bidirectional generation capabilities to construct high-quality reasoning traces when ground truth answers are available. When the policy model fails to generate high-quality outputs in a group (i.e., no correct math solution or no high-IoU bounding boxes), we manually insert the ground truth answer token to the end of a fully masked sequence and leverage dLLM's text-infilling capabilities to inpaint intermediate reasoning traces that lead towards the final answer. We name this guided generation approach \textbf{Answer Forcing}. An example is shown in Fig. \ref{fig:answer-forcing}.

Formally, given a prompt $\bm{x}$, we first sample group responses $\bm{y}^i,\cdots,\bm{y}^N\sim \pitheta(\bm{y}|\bm{x})$. Each $\bm{y}^i$ typically starts with a text reasoning trace enclosed by ``$\textlangle$ \texttt{think} $\textrangle $...$\textlangle$/ \texttt{think} $\textrangle $''  tags followed by the final answer enclosed in ``$\textlangle$ \texttt{answer} $\textrangle $...$\textlangle$/ \texttt{answer} $\textrangle $'' tags. The final answer can be either text or image tokens, depending on the tasks.

If all of the extracted answers have low rewards (e.g., incorrect for math reasoning tasks), and a ground truth answer $z^*$ is accessible, we can initialize a new sequence $\bm{y}^{N+1}=$ ``M \dots M $\textlangle$ \texttt{answer} $\textrangle $ $z^*$$\textlangle$/ \texttt{answer} $\textrangle $"  where M is the special mask token. We then employ the dLLM $\pi_\theta$ to progressively unmask these tokens and effectively generate a synthetic reasoning trace conditioned on the ground-truth answer. This sample is then added to the group.  Additional details of answer-forcing are included in Appendix \ref{sec:appendix_tree_search}

% This design is partially inspired by IGPO \cite{zhao2025inpainting}, which injects partially masked ground-truth reasoning traces as hint in the sampling process. Crucially, unlike IGPO which requires gold reasoning trace, Answer-Forcing does not require any pre-existing reasoning data, which are especially hard to obtain for data scarece tasks like object grounding. 

% Given a group size of $N$, the naive implementation would first generate $N$ samples, evaluate the rewards, and then decide whether to generate an additional sample via injection. This is highly inefficient. Instead, we always generate $N+1$ samples in parallel for each group, with 1 sample containing a ground-truth answer. However, depending on the rewards of the first $N$ samples, we optionally discard the extra sample from the loss computation when the remaining $N$ samples already include outputs with a significantly effective training signal (e.g., high rewards as measured by accuracy or IoU).

\subsubsection{Tree Search}
\begin{figure*}[t]
    \centering
    \includegraphics[width=0.85\linewidth]{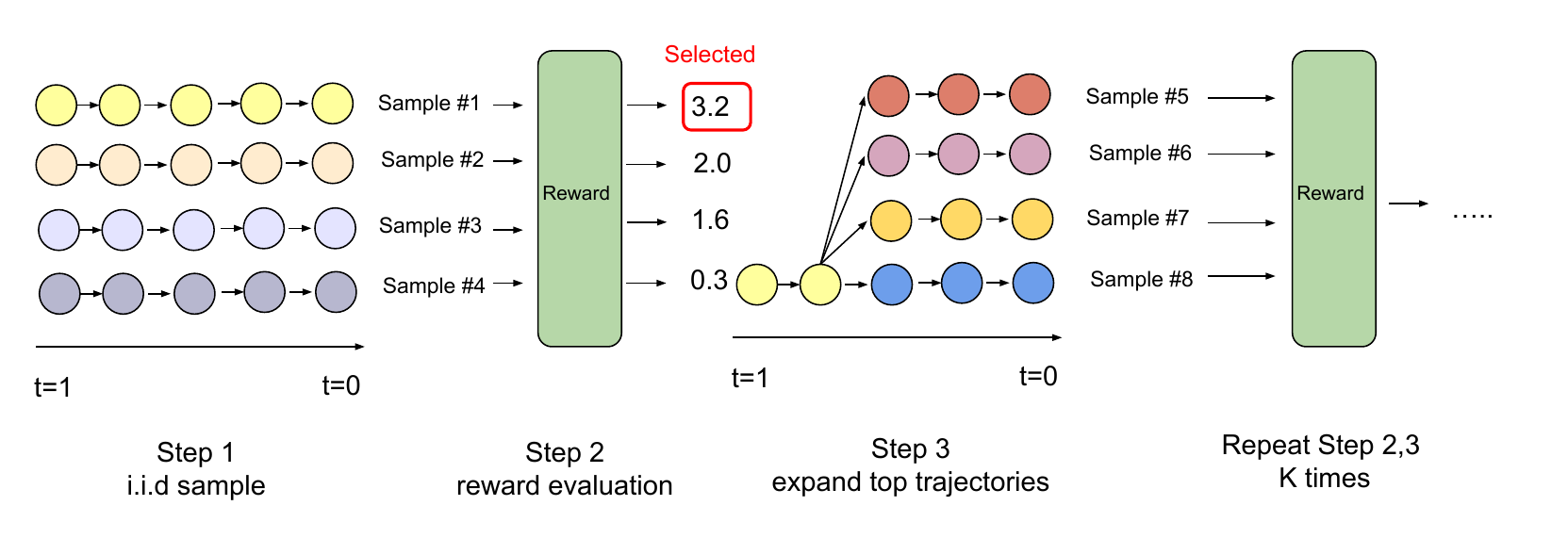}
    \caption{\textbf{Tree Search}. Given base group size $N$, we first sample $N$ i.i.d samples and evaluate the rewards. We then select the samples with the highest rewards and generate $N$ new samples from an early diffusion state of the best sample. This process is repeated $K$ times. In this example, $N=4$.}
    \label{fig:tree_search}
    \vspace{-1em}
\end{figure*}

For tasks that do not provide ground-truth answers (such as image editing), we leverage \textbf{Tree Search} to obtain high-reward rollouts. Given a base group size of $N$, we first generate $N$ samples and compute rewards as usual. We then find the samples in each group with the highest rewards and generate $N$ new samples, starting from an early state in the generated trajectories of those samples rather than from fully noised sequences. This gives $N$ new samples. This process is repeated $k$ times, yielding a final effective group size of $Nk$. This process is illustrated in Figure \ref{fig:tree_search}.

Concretely, given a prompt $\bm{x}$, we generate $N$ sequences $\bm{y}^1_0,\dots,\bm{y}^N_0$ through $T$ diffusion steps. We also keep track of intermediate diffusion states $\bm{y}^1_{t_i}, \dots, \bm{y}^N_{t_i}$ where $1=t_0>t_1>..t_T=0$ are discretized diffusion timesteps. Notably, $\bm{y}^1_{t_0}, \dots, \bm{y}^N_{t_0}$ are fully masked sequences and $\bm{y}^1_{t_T}, \dots, \bm{y}^N_{t_T}$ are final generated responses, which may contain both image and text tokens. After obtaining rewards $r_1, \cdots, r_N$ for each response, we find an index $m=\text{argmax}(r_1,...r_N)$ with the highest rewards and retrieve its early diffusion states $\bm{y}^m_{t_s}$, which is a partially masked sequence. The selection of the timestep $t_s\in\{t_0,...,t_T\}$, is controlled by a hyper-parameter. We then proceed to generate $N$ new samples $\bm{y}^{N+1}_0..y^{2N}_0$ using $\bm{y}^m_{t_s}$ as the initialization as opposed to a fully masked sequence. To generate these samples, we only need to perform $T-s$ diffusion steps. This process is repeated $k$ times until all $Nk$ samples are obtained. 

% Instead of saving a large number of trajectories in memory, we make use of the observation that once a token is unmasked, it will never be changed. Hence, we can return to an early state with only the final generated sequences and their token-generation ordering.  Given a trajectory $\bm{y}^j_{t_0}, \dots, \bm{y}^j_{t_T}$ for sample $\bm{y}^j_0$ with sequence length $L$, we only need to store the final generated result $\bm{y}^j_0=\bm{y}^j_{t_T}$ and an array $v$ of length $L$ to keep track of when each token is unmasked. $v[i]=p$ indicates the i-$th$ token in the sequence is unmasked at p-$th$ diffusion step where $1\le p \le T$. To recover  $\bm{y}^j_{t_s}$ for arbitrary $s$, we can simply use the relation $\bm{y}^j_{t_s}[i]=\bm{y}^j_0[i]$ if $v[i]\le s$ and $\bm{y}^j_{t_s}[i]=\M$  otherwise. This reduces the memory overhead of saving all trajectories and avoids the need to gather large trajectory data across GPUs when a group is distributed across multiple processes. 

\subsection{Complementary-Masking Likelihood Estimator}
\label{sec:log_prob}

One essential challenge in applying policy gradient methods to dLLMs is the estimation of the data log probability $\log\pi_\theta(\bm{y}|\bm{x})$. Unlike AR models whose likelihood has an exact computable form, dLLMs' likelihoods are estimated via the ELBO surrogate. The ELBO for the sequence log probability is expressed as the following formula,
\begin{align*}
\log \pi_\theta(\bm{y}|\bm{x}) =\mathbb{E}_{t, \bm{y}_t}\Big[w(t)\sum_{k\sim M(y_t)} \log \pitheta(\bm{y}[k] |\bm{y}_t, \bm{x})\Big]
\label{eq:estimator}
\end{align*}

where $w(t)$ is a weighting function and $M(y)=\{k|\bm{y}[k]=M\}$ is the set of masked indices. The expectation is typically computed via Monte Carlo (MC) Estimator. Existing works on dLLM RL mostly distinguish themselves from others through the choice of $w(t)$ and how they sample $\bm{y}_t$ in each MC sample. For example, d1 \cite{zhao2025d1} samples one MC sample at $t = 1$ (i.e. fully-masked sequence), and adopts the weighting $w(t) = 1/t$; UniGRPO \cite{yang2025mmada} samples one MC sample at $t\sim \text{Uniform}([0,1])$ with $w(t) = 1/t$.

In our design, we use two samples with timestep $t_1\sim \text{Uniform}([0,1])$ and $t_2=1-t$. We sample $\bm{y}_{t_1}\sim q(y_{t_1}|y_0)$ using the discrete forward diffusion process, and set $\bm{y}_{t_2}[i]=\bm{y}[i]$ if $\bm{y}_{t_1}[i]=M$ and  $\bm{y}_{t_2}[i]=M$  if $\bm{y}_{t_1}[i]\ne M$. This design, known as complementary masking, was first proposed in LaViDa \cite{li2025lavida} for pretraining. For example, if the sequence is $
\bm{y}$ is ``\texttt{there is a dog}" and $
\bm{y}_{t_1}$ is ``\texttt{[M] is [M] dog}", $\bm{y}_{t_2}$ will be ``\texttt{there [M] a [M]}".  We adopt a $w(t)=1$ instead of  $w(t)=\frac{1}{t}$ from LaViDa, giving the following estimator
\begin{align*}
\log \pitheta(\bm{y}|\bm{x}) = \frac{1}{2}\sum_{j \in\{1, 2\}}\sum_{k\sim M(y_{t_j})} \log \pitheta(\bm{y}[k]|\bm{y}_{t_j}, \bm{x})
\end{align*}

%  DMPO \cite{zhu2025enhancing}

% DMPO \cite{zhu2025enhancing} generates four mask samples for each sequence, forming two pairs of complimentary masked sequences, and still adopts the weighting $w(t) = 1/t$. While these estimations are good enough for achieving satisfactory performances on single-modality tasks, we found that they remain suboptimal and, more importantly, prone to numerical instability due to the choice of an exploding weighting $w(t) = 1/t$. To address this issue, we propose the following simple log-probability estimation recipe that is robust across multiple modalities in RL tasks while remaining efficient. We illustrate the computation and its difference from other methods in Figure \ref{fig:likelihood}.

% For each clean response sequence $\bm{y}$ of length $L$, we sample two mask samples $\bm{y}_{t_1}, \bm{y}_{t_2}$ for ELBO computation using the following protocol: we first choose uniformly an integer $k$ between $0$ and $L$, then we select a set of indices $\mathcal{M}$ in $\bm{y}$ at random, consisting of exactly $k$ position, to turn into mask $[M]$, creating the first mask sample $\bm{y_{t_1}}$. Then, we turn the rest of indices $\{0, \dots, L\} \setminus \mathcal{M}$ in $\bm{y}$ into mask, creating a second mask sample $\bm{y}_{t_2}$ with a coupled, complimentary mask as $\bm{y}_{t_1}$. Finally, we compute the ELBO through the following formula with a uniform weight $w(t) = 1$,

Our estimation recipe has several advantages. First, compare with i.i.d MC samples, it masks all tokens once, ensuring the estimate accounts for all tokens in the sequence. This prevents important tokens from being disregarded during training. Second, compare with d1, which can also get estimates for all tokens by always masking every token, it has a smaller training-inference gap. Finally, compared with naively applying complementary masking with $w(t)=\frac{1}{t}$, using $w(t)=1$ avoids imbalanced token weighting caused by drastically different masking ratios. When $w(t)=\frac{1}{t}$, suppose $t_1=0.9$ and $t_2=0.1$, we have $\frac{w(t_2)}{w(t_1)}=9$, indicating that tokens in sample $y_{t_2}$ is $9\times$ more important than those in sample $y_{t_1}$, which is highly unideal since which tokens are masked in which sample is randomly determined.

% Our choice of complementarily masked samples introduces an additional coupling, ensuring that the model's outputs at each token position are included in the computation and enhancing the supervision signal. Moreover, we use a simple uniform weighting $w(t) = 1$, avoiding distributing overly imbalanced weights over sequences with drastically different mask ratios. This achieves a great numerical stability for the RL algorithm, which is critical for a multi-task, multi-modality learning setup. Concurrent work \cite{shi2025demystifying} has also found that such a uniformly weighted ELBO is technically tighter and induces superior performances across many other weighting function choices. These designs, combined, yield a robust log-likelihood estimator, enabling us to perform online RL training effectively. 

\section{Experiments}

\subsection{Setup}

We select LaViDa-O as our base model because of its strong multimodal performances and pre-existing reasoning capabilities \cite{li2025lavidao}. \ours~involves two training stages: the first stage is supervised finetuning (SFT) on reasoning data, the second stage is unified post-training on a mix of SFT and RL data using a mix SFT, RL, and self-distillation loss under our unified framework. The RL datasets consist of math reasoning, visual question answering, reason-intensive object grounding and image editing. We use correctness reward for math and QA problems, IoU rewards for object grounding, and the EditScore \cite{luo2025editscore} reward model for image editing. We defer further details on the dataset composition, training schedule, and hyperparameter to Appendix \ref{sec:appendix_setup}.

\subsection{Image Understanding Results}
\begin{table*}[t]
\centering
\small
\caption{\textbf{Performance comparison across visual reasoning, VQA, and language-only benchmarks.} *per-dataset finetuning results. $^\dagger$ RL checkpoint for MMada is not open-sourced. Original authors only reported a limited set of results.}
\label{tab:result_und}
\scriptsize
\setlength{\tabcolsep}{11pt} % Adjust this value to control the width
\begin{tabular}{lccccccc}
\toprule
 & \multicolumn{2}{c}{\textbf{Visual Reasoning}} 
 & \multicolumn{3}{c}{\textbf{Visual QA}} 
 & \multicolumn{2}{c}{\textbf{Text Only}} \\
\cmidrule(lr){2-3} \cmidrule(lr){4-6} \cmidrule(lr){7-8}
\textbf{Model} 
& MathVista & MathVerse 
& ChartQA & AI2D & MMMU-Pro
& GSM8K & MATH-500 \\
\midrule
\multicolumn{8}{c}{\textit{Language-Only dLLMs}} \\
LLaDA-8B-Instruct \cite{nie2025large}& --  & --&  --  &  --  & --   & 78.2    &  36.2 \\ 
+ DiffuGRPO \cite{zhao2025d1}  & --  & --&  --  &  --  & --   & 82.1$^*$    &  40.2$^*$  \\ 
Dream-7B\cite{dream2025}    & --&  --  &  --  & --   & - & 77.2 & 39.6\\ 

\midrule
\multicolumn{8}{c}{\textit{Visual-Understanding-Only dLLMs}} \\
LaViDa-L \cite{li2025lavida}      & 44.8 & 27.2 &64.6  & 70.0   & 27.1  & --    & --   \\
Dimple \cite{yu2025dimple}   & 42.3 & -- & 63.4  & 74.4  & --    & --  & --   \\
% LLaDa-V      & -- & -- & --   & --   & --   & --    & --   \\
\midrule
\multicolumn{8}{c}{\textit{Unified-Understanding-and-Generation dLLMs}} \\
MMaDa-8B-Base   \cite{yang2025mmada}   & 27.1 & 13.4 &  9.6  &  56.1  & 3.2   & 17.4    &  4.2  \\
+CoT SFT \cite{yang2025mmada}   &33.7 & 13.5 & 9.8   & 66.6  & 8.4   & 65.2 & 26.5 \\
+UniGRPO$^\dagger$  \cite{yang2025mmada} & -- & -- & --   & --   & --   & 73.4 &  36.0 \\
\hline
LaViDa-O \cite{li2025lavidao}      & 56.9 & 36.9 & 80.0 & 76.7 & 31.2 & 47.4 & 23.4 \\
\rowcolor{gray!20} 
+SFT    & 57.6 & 36.6 & 80.8 & 78.6 & 31.9 & 70.6  & 31.0 \\
% LaViDa-GRPO   & --   & --   & --   & --   & --   & --    & --   \\
\rowcolor{gray!20} 
LaViDa-R1     & \textbf{60.0} & \textbf{38.7} & \textbf{81.7} & \textbf{78.9} & \textbf{32.8} & \textbf{81.5} & \textbf{38.6} \\
\bottomrule
\end{tabular}

\end{table*}

\begin{table*}[h!]
\centering
\caption{\textbf{Per-Category and overall scores on ImgEdit benchmark.}}
\label{tab:image-edit}
%\resizebox{1.0\linewidth}{!}{ 
\scriptsize
\setlength{\tabcolsep}{7pt} % Adjust this value to control the width
{
\begin{tabular}{lcccccccccc}
\hline
\textbf{Model} & \textbf{Add} & \textbf{Adjust} & \textbf{Extract} & \textbf{Replace} & \textbf{Remove} & \textbf{Background} & \textbf{Style} & \textbf{Hybrid} & \textbf{Action} & \textbf{Overall} \\
\hline
GPT-4o \citep{openai2024gpt4o} & 4.61 & 4.33 & 2.90 & 4.35 & 3.66 & 4.57 & 4.93 & 3.96 & 4.89 & 4.20 \\
% Ovis-U1 \citep{wang2025ovis} & 4.13 & 3.62 & 2.98 & 4.45 & 4.06 & 4.22 & 4.69 & 3.45 & 4.61 & 4.00 \\
Qwen2.5VL+Flux \citep{wang2025gpt} & 4.07 & 3.79 & 2.04 & 4.13 & 3.89 & 3.90 & 4.84 & 3.04 & 4.52 & 3.80 \\
% \hline

% \hline

% LaViDa-O & \textbf{3.86} & \textbf{3.39} & 2.26 & \textbf{4.16} & \textbf{3.82} & \textbf{3.79} & \textbf{4.81} & 2.25 & 3.96 & \textbf{3.59} \\
% Ours-newdata-crop-30k & 3.71 & 3.53 & 2.15 & 4.03 & 3.54 & 3.76 & 4.71 & 2.72 & 3.82 & 3.55 \\
FluxKontext dev \citep{labs2025flux1kontextflowmatching} & 3.76 & 3.45 & 2.15 & 3.98 & 2.94 & 3.78 & 4.38 & 2.96 & 4.26 & 3.52 \\
OmniGen2 \citep{wu2025omnigen2}& 3.57 & 3.06 & 1.77 & 3.74 & 3.20 & 3.57 & 4.81 & 2.52 & 4.68 & 3.44 \\
UniWorld-V1 \citep{lin2025uniworld} & 3.82 & 3.64 & 2.27 & 3.47 & 3.24 & 2.99 & 4.21 & 2.96 & 2.74 & 3.26 \\
BAGEL \citep{deng2025emerging}& 3.56 & 3.31 & 1.70 & 3.30 & 2.62 & 3.24 & 4.49 & 2.38 & 4.17 & 3.20 \\
Step1X-Edit \citep{liu2025step1x}  & 3.88 & 3.14 & 1.76 & 3.40 & 2.41 & 3.16 & 4.63 & 2.64 & 2.52 & 3.06 \\
% ICEdit \citep{zhang2025ice} & 3.58 & 3.39 & 1.73 & 3.15 & 2.93 & 3.08 & 3.84 & 2.04 & 3.68 & 3.05 \\
OmniGen \citep{xiao2025omnigen1} & 3.47 & 3.04 & 1.71 & 2.94 & 2.43 & 3.21 & 4.19 & 2.24 & 3.38 & 2.96 \\
UltraEdit \citep{zhao2024ultraedit} & 3.44 & 2.81 & 2.13 & 2.96 & 1.45 & 2.83 & 3.76 & 1.91 & 2.98 & 2.70 \\
AnyEdit \citep{yu2025anyedit} & 3.18 & 2.95 & 1.88 & 2.47 & 2.23 & 2.24 & 2.85 & 1.56 & 2.65 & 2.45 \\
InstructAny2Pix\citep{li2023instructany2pix} & 2.55 & 1.83 & 2.10 & 2.54 & 1.17 & 2.01 & 3.51 & 1.42 & 1.98 & 2.12 \\ 
MagicBrush \citep{zhang2023magicbrush} & 2.84 & 1.58 & 1.51 & 1.97 & 1.58 & 1.75 & 2.38 & 1.62 & 1.22 & 1.90 \\
Instruct-Pix2Pix\citep{brooks2023instructpix2pix} & 2.45 & 1.83 & 1.44 & 2.01 & 1.50 & 1.44 & 3.55 & 1.20 & 1.46 & 1.88 \\
\hline
% \rowcolor{gray!20}
LaViDa-O \citep{li2025lavidao} & 4.04	&3.62	&2.01&	4.39	&3.98	&\textbf{4.06}	&4.82&	2.94 &	3.54&	3.71 \\
+ Reasoning & 4.11	& 3.67&	2.04	& 4.40&	 4.05	&4.00&	4.75 &	3.10	&4.04&	3.80 \\
\rowcolor{gray!20} 
+ SFT & 4.11&		3.80&		2.21&		4.46& 3.90	&	3.86	&	4.76	&	3.09	&	4.14	&	3.81 \\
\rowcolor{gray!20} 

% LaViDa-R1 & \textbf{4.20} &\textbf{	3.85}&	\textbf{2.33}	&4.34&	3.98&	4.00 &	\textbf{4.86}&	\textbf{3.14}	&\textbf{4.19}&	\textbf{3.88} \\

LaViDa-R1 & \textbf{4.25} &\textbf{3.90}&	\textbf{2.32}	&\textbf{4.52}&	\textbf{4.06} &	3.86 &	\textbf{4.87}&	\textbf{3.10}	&\textbf{4.18}&	\textbf{3.90} \\

% + Planning & 4.11	& 3.67&	2.04	& 4.40&	 4.05	&4.00&	4.75 &	3.10	&4.04&	3.80 \\
\hline
\end{tabular}
}

\end{table*}

We report results on a wide range of visual understanding tasks and language-only tasks in Table \ref{tab:result_und}. We report results on MathVista \cite{lu2023mathvista} and MathVerse \cite{zhang2024mathverse} for visual math reasoning, ChartQA, AI2D and MMMU-Pro \cite{masry-etal-2022-chartqa,kembhavi2016diagram,yue2025mmmu} for visual QA and GSM8K and Math500 \cite{cobbe2021gsm8k,lightman2023lets} for language-only tasks. For all datasets, we report the accuracy metric. Notably, we observe that \ours~show improvements across all tasks, with the biggest gain coming from the language-only GSM8K and Math500 datasets. We hypothesize that this is because the base model's pretraining dataset is vision-centric, leading to poor language performance and leaving considerable room for improvement.

To further validate the effectiveness of \ours, we evaluate on additional multimodal understanding benchmarks that are less reason-intensive such as MMMU \cite{yue2023mmmu}, MMBench \cite{MMBench}, MME \cite{fu2023mme} and report results in Table \ref{tab:additional_und_benchmarks}. Results show that \ours~leads to consistent improves. We provide additional qualitative results in Appendix \ref{sec:appendix_qualitative_results}.

\begin{table}[t]
\centering
\caption{\textbf{Additional multimodal benchmark results.}}
\label{tab:additional_und_benchmarks}
\scriptsize
\begin{tabular}{lcccc}
\toprule
\textbf{Model} & \textbf{Capabilities} & \textbf{MMMU} & \textbf{MMB} & \textbf{MME} \\
\midrule
LLaDa-V   & Understanding Only   & 48.6 & 82.9 & 491 \\
MMaDa     & Und \& Gen & 30.2 & 68.5 & 242 \\
\rowcolor{gray!20} 
LaViDa-O  & Und \& Gen & 45.1 & 76.4 & 488 \\
\rowcolor{gray!20} 
LaViDa-R1 & Und \& Gen & \textbf{47.0}  & \textbf{79.2} & \textbf{501}  \\
\bottomrule
\end{tabular}
\end{table}

\subsection{Image Editing Results}

We evaluate image editing performance on the ImgEdit benchmark~\cite{ye2025imgedit}, and report the benchmark scores in Table \ref{tab:image-edit}. These scores measure both visual quality and prompt compliance via a GPT-4 judge model. We note that the base model LaViDa-O already included some reasoning data in its training pipeline, and has reported image editing performance with reasoning. %Compared with this baseline, SFT only leads to a marginal improvement (+0.01), suggesting the base model is already well-trained.  However, additional post-training with proposed unified algorithm leads to considerable additional gain (+0.10), highlighting the effectiveness of proposed algorithm.
While SFT leads to a negligible improvement (+0.01) , indicating a performance saturation characteristic of supervised scaling, LaViDa-R1 achieves a significant boost (+0.10). This underscores that our unified RL framework successfully drives exploration beyond the modes learned during supervision. We provide additional qualitative results in Appendix \ref{sec:appendix_qualitative_results}.
% \ours~achieves higher accuracy (+0.08) compared to LaViDa-O and other state-of-the-art unified models such as BAGEL~\cite{deng2025emerging}, while reducing end-to-end latency from 63.98s to 22.55s, achieving a $2.83\times$ speedup.

\subsection{Reason-Intensive Grounding}

\begin{table}[t]
\centering
\small
\caption{\textbf{Performance comparison on Lisa-Grounding Dataset.}}
\label{tab:grounding}
\begin{tabular}{lcc}
\toprule
\textbf{Model} & \textbf{P@0.5} & \textbf{mIoU$_\text{box}$} \\
\midrule
\multicolumn{3}{c}{\textit{Specialist Models}} \\
SegLLM \cite{wang2025segllm} & 61.3 & 55.2   \\
LISA-7B \cite{lai2024lisa} & 49.4 & 50.6   \\
\midrule
\multicolumn{3}{c}{\textit{General-purpose VLMs}} \\
Qwen2.5-VL-7B \cite{bai2025qwen25-vl} & 32.0 & 28.7   \\
Qwen3-VL-8B\cite{Qwen3-VL} & 62.4 & 56.6   \\
% InternVL3-8B-Instruct & --& --   \\\\
\midrule
\multicolumn{3}{c}{\textit{Reinforcement Learning}} \\
VLM-R1 \cite{shen2025vlm}         & 63.1 & --   \\
\hline
LaViDa-O \cite{li2025lavidao} & 29.2  & 26.1 \\ 
\rowcolor{gray!20} 
+SFT    & 40.3  & 36.9 \\
\rowcolor{gray!20} 
LaViDa-R1       & \textbf{66.7}  & \textbf{60.0} \\

\hline
\end{tabular}

\end{table}
We evaluate reason-intensive grounding on Lisa-Grounding dataset\cite{lai2024lisa}, and report results in Table \ref{tab:grounding}.  We report precision@0.5 (P@0.5) and the mean IoU (mIoU) of bounding boxes. While the base model LaViDa-O exhibits strong grounding performance on simple queries, it performs poorly on Lisa-Grounding which requires complex visual reasoning. Compared with this baseline, SFT improves the performance by $+10.8$ mIoU and unified post-training further improves the performance by an additional $+22.1$ mIoU. We provide additional qualitative results in Appendix \ref{sec:appendix_qualitative_results}.

\section{Ablation Studies}

To verify the effectiveness of \ours, we conduct additional ablation studies to assess several design choices. 

\textbf{Answer Forcing.} We investigate the effectiveness of answer forcing and report results in Table \ref{tab:ablation_injection}. We explored answer forcing randomly with probabilities of 0\%, 10\%, 50\%, and 100\%. Results demonstrate that 10\% forcing has the best overall performance. A high inject ratio leads to collapse because answer-forced samples always receive a high correctness reward, even when their reasoning traces are ill-formed, thereby producing potentially misleading learning signals. This problem is particularly severe when answer forcing always occurs, since it implies that most other samples will likely have a negative advantage due to centering. 

\begin{table}[t]
    \centering
        \caption{\textbf{Ablation Studies of Answer-Forcing.} *Collapsed}
    \label{tab:ablation_injection}
\begin{tabular}{lccc}
\toprule
 & \textbf{M.Vista} & \textbf{Lisa-Gnd.} & \textbf{Math500} \\
\midrule
Inject 0\%   & 57.8  & 63.1 & 36.2 \\
Inject 10\%  & \textbf{58.9}  & \textbf{65.0} & \textbf{38.0} \\
Inject 50\%  & 58.0  & 64.2 & 35.4 \\
Inject 100\% & 4.1*  & 5.1* & 4.2* \\
\bottomrule
\end{tabular}

\end{table}
\begin{table}[t]
    \centering
        \caption{\textbf{Ablation Studies of Tree search}}
    \label{tab:ablation_tree_search}
\begin{tabular}{lcc}
\toprule
\textbf{Tree Search Steps} & \textbf{Group Size} & \textbf{ImgEdit} \\
\midrule
N/A              & 16      & 3.85 \\
N/A              & 32      & 3.84 \\
N/A             & 64      & 3.84 \\
$[0,8]$         & $16\times 2$ & \textbf{3.90} \\
$[0,8,16,32]$   & $16\times 4$ & 3.87 \\
\bottomrule
\end{tabular}

\end{table}
\textbf{Tree Search.} The hyperparameter that controls the tree search behavior is called restart timestep indices, which is a list of integers specifying the branching steps. For example, given a group size of 16 samples per prompt and a 64-step generation pipeline,  a tree search of [0,8] means we first sample 16 outputs independently, each run for 64 steps. We then identify the trajectory corresponding to the sample with the highest reward and branch from its 8th step to generate 16 additional samples. These 16 new samples are initialized from the 8th step of the best-performing sample previously generated, underwriting 56 steps each. Results are shown in Tab. \ref{tab:ablation_tree_search}. Steps [0,8] is a good choice. Steps [0, 8, 16, 32] yield almost identical performance, because starting from a later diffusion step introduces less uncertainty and does not contribute much.

\begin{table}[t]
    \centering
        \caption{\textbf{Ablation Studies of Likelyhood Estimator}}
    \label{tab:ablation_liklihood}
\begin{tabular}{c cHH c c}
\toprule
\textbf{$\#$MC} & \textbf{Masking} & \textbf{Source} & \textbf{Math500} & \textbf{Lisa-Grounding} & \textbf{ImgEdit} \\
\midrule
1 & i.i.d           & UniGRPO     & TODO & 61.9 & 3.82 \\
1 & Full    & d1          & TODO & 59.2 & 3.77 \\
2 & i.i.d           & UniGRPO     & TODO & 62.1 & 3.86 \\
2 & Compl.   & LaViDa-R1   & 38   & \textbf{65.0} & \textbf{3.88} \\
\bottomrule
\end{tabular}

\end{table}
\textbf{Likelihood Estimator.} We investigate the effectiveness of our simple likelihood estimation recipe and report results in Table \ref{tab:ablation_liklihood}. We explored four approaches. In the first setup (row 1), we randomly mask a subset of tokens and compute the likelihood only over masked positions. This is equivalent to UniGRPO with 1 MC sample. In the second setup (row 2), we mask all tokens and thus compute the likelihood over all tokens, which is equivalent to the d1 setup. In the third setup, 2 i.i.d. MC mask samples were explored. Finally, we report the results of our proposed estimation recipe (row 4). The results show that our method achieves the best performance.

\textbf{Self-Distillation Loss.} We experimented with varying $\gamma$,the weight of self-distillation loss described in Section \ref{sec:method_unified_training} and report results in Table \ref{tab:ablation_loss}. The results show that combining two loss functions yields better performance. Intuitively, this loss assigns greater importance to the best-generated samples than standard GRPO. These results highlight the flexibility of the proposed unified paradigm. 

\begin{table}[h]
    \centering
        \caption{\textbf{Ablation Studies of Self-Distillation Loss}}
          \label{tab:ablation_loss}
\begin{tabular}{lcc}
\toprule
\textbf{$\gamma$} & Effective Objective &ImgEdit \\
\midrule
$\gamma = 0$   & On-Policy GRPO  & 3.86 \\
$\gamma = 0.5$ & Mixed & \textbf{3.90} \\
$\gamma = 1.0$ &  Self-Distillation & 3.84 \\
\bottomrule
\end{tabular}
\end{table}

\begin{figure}[h]
    \centering
    \includegraphics[width=0.95\linewidth]{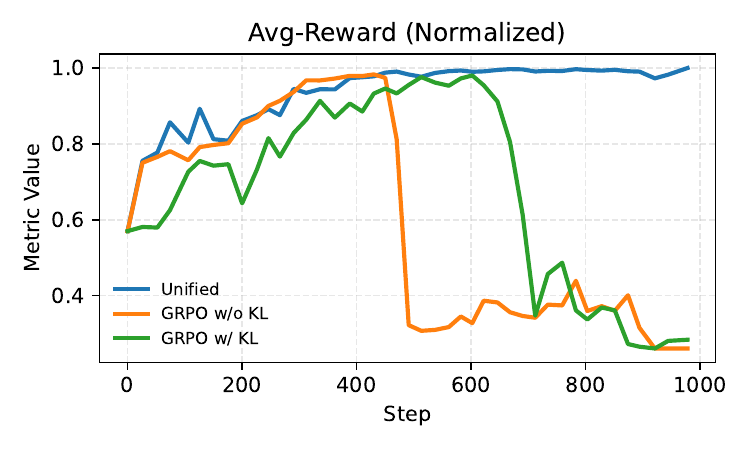}
    \caption{\textbf{Ablation Studies of Unified Objective.}}
    \label{fig:ablation_unified}
    \vspace{-1em}
\end{figure}

\textbf{Unified Loss.} We finally investigated the effectiveness of the proposed unified loss that combines multiple objectives. We plot the average reward per sample during training. The results are shown in Figure \ref{fig:ablation_unified}. We compare with the standard online GRPO with and without the KL regularizer. Results show that the proposed unified loss with SFT as a regularization term is more stable and yields higher reward. We observe that GRPO diverges even with strong KL regularization $\beta=0.1$ because the KL estimators only compute divergences on sampled tokens and are not suitable for high-entropy image distributions. Specifically, for most samples, the negative-log-likelihood is above 6 for visual generation and is less than 2 for text generation. The high NLL leads to high variance in the KL term.

\section{Conclusion}
In this paper, we propose \ours, a novel training recipe to enhance reasoning ability in unified multimodal dLLMs. LaViDa-R1 introduces a unified post-training paradigm via weighted policy-gradient objectives and a simple yet effective likelihood estimator for stable training. LaViDa-R1 also adopts two new guided rollout generation algorithms to address the key issue of vanishing training signal in online RL. Through multi-task, multi-reward, cross-modality RL, LaViDa-R1 achieves superior performances across a wide range of tasks, including text-only and multimodal reasoning, visual QA, image grounding, and editing.

\section*{Impact Statement}
This paper presents work aimed at advancing the field of machine learning. It proposes a unified multimodal model capable of generating text and images, thereby inheriting the full potential of LLMs and image generators. For example, it may be abused to create various harmful and offensive content. We strongly caution the community against such use cases.

\section*{Acknowledgements}
AG was supported by NSF CAREER Grant \#2341040 and a Schmidt AI 2050 Fellowship. MT is grateful for partial support by NSF Grant DMS-2513699, DOE Grants NA0004261, SC0026274, Richard Duke Fellowship, and Simons Institute for the Theory of Computing at UC Berkeley. YC is grateful for the support from NSF Grant DMS-2450378.

% Authors are \textbf{required} to include a statement of the potential broader
% impact of their work, including its ethical aspects and future societal
% consequences. This statement should be in an unnumbered section at the end of
% the paper (co-located with Acknowledgements -- the two may appear in either
% order, but both must be before References), and does not count toward the paper
% page limit. In many cases, where the ethical impacts and expected societal
% implications are those that are well established when advancing the field of
% Machine Learning, substantial discussion is not required, and a simple
% statement such as the following will suffice:

% ``This paper presents work whose goal is to advance the field of Machine
% Learning. There are many potential societal consequences of our work, none
% which we feel must be specifically highlighted here.''

% The above statement can be used verbatim in such cases, but we encourage
% authors to think about whether there is content which does warrant further
% discussion, as this statement will be apparent if the paper is later flagged
% for ethics review.

% In the unusual situation where you want a paper to appear in the
% references without citing it in the main text, use \nocite
% \nocite{langley00}

\bibliography{ref}
\bibliographystyle{icml2026}

%%%%%%%%%%%%%%%%%%%%%%%%%%%%%%%%%%%%%%%%%%%%%%%%%%%%%%%%%%%%%%%%%%%%%%%%%%%%%%%
%%%%%%%%%%%%%%%%%%%%%%%%%%%%%%%%%%%%%%%%%%%%%%%%%%%%%%%%%%%%%%%%%%%%%%%%%%%%%%%
% APPENDIX
%%%%%%%%%%%%%%%%%%%%%%%%%%%%%%%%%%%%%%%%%%%%%%%%%%%%%%%%%%%%%%%%%%%%%%%%%%%%%%%
%%%%%%%%%%%%%%%%%%%%%%%%%%%%%%%%%%%%%%%%%%%%%%%%%%%%%%%%%%%%%%%%%%%%%%%%%%%%%%%
\newpage
\appendix
\onecolumn

\section{Additional Technical Details}
\subsection{Formulation of dLLM}
\newcommand{\cat}[0]{\text{Cat}}
\newcommand{\alphats}[0]{\frac{1-t}{1-s}}
\newcommand{\oneminusalphats}[0]{\frac{t-s}{1-s}}

In this section, we provide an overview of the standard formulation of dLLMs that are adopted by the literature \cite{ou2024your,shi2024simplified,sahoo2024simple,lou2023discrete-sedd,you2025lladav,li2025lavida,li2025lavidao}. Notations are adapted from these references to be consistent with the ones used in the main paper to avoid potential confusion. 

Given a sequence of discrete tokens $\bm{y}_0$ whose lengths is $L$, the forward discrete diffusion process $q(\bm{y}_t|\bm{y}_s)$ gradually replace the original tokens in $\bm{y}_0$ to a special mask token $[M]$ over the time interval $[0,1]$, with $1 \ge t \ge s \ge 0$. At $t=1$, the sequence $\bm{y}_1$ is a fully masked sequence. This forward process is formally defined as

% Each token $X_t^i$ belongs to a fixed-size vocabulary set $V$. In our setup, $V$ consists of text tokens, image VQ tokens, and the special mask token $[M]$. This forward process is formally defined as

\begin{equation}
    q(\bm{y}_t^i|\bm{y}_s^i) =  
    \begin{cases}
      \cat(\bm{y}_t[i];\textbf{M}), & \text{if } \bm{y}_s[i]=[M] \\
      \cat(\bm{y}_t[i];\alphats \mathbf{Y_s}[i]+\oneminusalphats \textbf{M}), & \text{if } \bm{y}_s[i] \ne [M],
    \end{cases}
\end{equation}

where $\cat(\cdot)$ denotes a discrete categorical distribution, and $\textbf{M}, \mathbf{Y_s[i]} \in \mathbb{R}^{|V|}$ are probability vectors, and $|V|$ is the vocabulary size. Specifically, $\textbf{M}$ is a one-hot vector corresponding to the special token $[M]$, and $\mathbf{Y_s[i]}$ is a one-hot vector corresponding to the token $\bm{y}_s^i$. It has been shown that this forward process has the following marginal distribution:

\begin{equation}
    q(\bm{y}_t[i]|\bm{y}_0[i]) =  \cat(\bm{y}_t[i];(1-t) \mathbf{Y_0[i]}+t \textbf{M}).
    \label{eq:q_process}
\end{equation}

MDLM \citep{sahoo2024simple} shows that the posterior of the reverse process $p(\bm{y}_s|\bm{y}_t,\bm{y}_0)$ has the following form:

\begin{equation}
    p(\bm{y}_s[i]|\bm{y}_t[i],\bm{y}_0[i]) =  
    \begin{cases}
      \cat(\bm{y}_s[i];\mathbf{Y_t}[i]), & \text{if } \bm{y}_s[i] \ne [M] \\
      \cat(\bm{y}_s[i];\tfrac{t-s}{t} \mathbf{Y_0}[i]+\tfrac{s}{t} \textbf{M}), & \text{if } \bm{y}_s[i] = [M].
    \end{cases}
    \label{eq:appendix-eq-p}
\end{equation}

In inference, the clean sequence $\bm{y}_0$ is not known at start, so it is replaced with the prediction from a policy network $\pi_\theta(\bm{y}_0[i]|\bm{y}_t)$, which gives the following empirical sampling process:

\begin{equation}
    p_\theta(\bm{y}_s[i]|\bm{y}_t) =  
    \begin{cases}
      \cat(X_s^i;\mathbf{\bm{Y}_t}[i]), & \text{if } \bm{y}_s[i] \ne [M] \\
      \cat(X_s^i;\tfrac{t-s}{t} \pi_\theta(\bm{y}_0[i]|\bm{y}_t)+\tfrac{s}{t} \textbf{M}), & \text{if } X_s^i = [M].
    \end{cases}
    \label{eq:appendix-inference}
\end{equation}

\textbf{SFT training process.} During SFT training, the following maximum likelihood estimation objective is adopted:

\begin{equation}
\mathcal{L}_{\text{SFT}}=-\mathbb{E}_{(\bm{y},\bm{x})\sim\mathcal{D}} [\log \pi_\theta(\bm{y}|\bm{x})]
    \label{eq:mle}
\end{equation}
where the likelihood is estimated via an MC estimator described in Sec. \ref{sec:log_prob} of the main paper. We provide further discussion of estimating the likelihood in Appendix \ref{sec:likelihood}.

% \textbf{Sampling process.} During sampling, we start with a fully masked sequence $\bm{y}_1$. We then discretize the continuous time interval $[0,1]$ into discrete timesteps $0=t_0<t_1<\cdots<t_K=1$, and iteratively sample $\bm{y}_{t_{k-1}}\sim p_\theta(\bm{y}_{t_{k-1}}|\bm{y}_{t_k})$ using Equation \ref{eq:appendix-inference}. We start with $k=K$ and end when we obtain a mask-free sequence $X_0$. At each step, we assume $\pi_\theta(\bm{y}_{t_{k-1}}|\bm{y}_{t_k})=\prod_{i=1}^L \pi_\theta(\bm{y}_{t_{k-1}}[i]|\bm{y}_{t_k})$, following the independence assumption of previous works \citep{nie2025large, sahoo2024simple, lou2023discrete-sedd}. 

% \textbf{Training process.} At each training step, given a clean sequence $X_0$, we sample a random timestep $t\in[0,1]$ and obtain $X_t\sim q(X_t|X_0)$ through the forward process defined in Equation \ref{eq:q_process}. This gives us a partially masked sequence. The loss is then computed using Equation \ref{eq:dlm-obj-ref} from Section \ref{sec:mdm_related}.

\subsection{Unified paradigm for post-training}
\label{sec:appendix_unified}
In this section, we provide a theoretical justification for unifying various post-training objectives into a weighted policy gradient method. We will include the derivation for GRPO \cite{guo2025deepseek}, Online DPO and its variants \cite{guo2024direct, zhao2023slic}, SFT, Best-of-N self-distillation \cite{sessa2024bond}. We will show that these objectives share the same gradients as certain policy-gradient objectives, with a special weight.

\paragraph{GRPO} We consider the setup of \textbf{fully-online} GRPO with strength of KL regularization $\beta = 0$. In this case, the GRPO objective is simplified to 
\begin{align*}
& J_{\text{grpo}}(\theta) = \underset{\bm{x}^i \sim \piold}{\mathbb{E}}\Big[\frac{1}{N} \sum_{i = 1}^{N} \min\Big(\dfrac{\pitheta(\bm{y}^i|\bm{x})}{\piold(\bm{y}^i|\bm{x})}A_i^{\text{GRPO}}, \operatorname{clip}\big(\dfrac{\pitheta(\bm{y}^i|\bm{x})}{\piold(\bm{y}^i|\bm{x})}, 1-\varepsilon, 1 + \varepsilon\big) A^{\text{GRPO}}_i\Big) \Big]
\end{align*}
In a pure on-policy setup, $\piold(\bm{y}^i | \bm{x}) = \operatorname{sg}(\pitheta)(\bm{y}^i|\bm{x})$. Using the fact that $\nabla \pitheta(\bm{y}^i|\bm{x}) = \operatorname{sg}(\pitheta)(\bm{y}^i|\bm{x}) \nabla \log \pitheta(\bm{y}^i|\bm{x})$, we can express the gradient of the GRPO objective as,
\begin{align*}
\nabla J_{\text{grpo}}(\theta) & = \underset{\bm{x}^i \sim \pitheta}{\mathbb{E}}\Big[\frac{1}{N} \sum_{i = 1}^{N} \min\Big(\dfrac{\operatorname{sg}(\pitheta(\bm{y}^i|\bm{x})}{\operatorname{sg}(\pitheta(\bm{y}^i|\bm{x})}A_i^{\text{GRPO}}, \operatorname{clip}\big(\dfrac{\operatorname{sg}(\pitheta(\bm{y}^i|\bm{x})}{\operatorname{sg}(\pitheta(\bm{y}^i|\bm{x})}, 1-\varepsilon, 1 + \varepsilon\big) A^{\text{GRPO}}_i \Big)  \nabla \log \pitheta(\bm{y}^i|\bm{x}) \Big]\\
& = \underset{\bm{x}^i \sim \pitheta}{\mathbb{E}}\Big[\frac{1}{N} \sum_{i = 1}^{N} \min\Big(A_i^{\text{GRPO}}, \operatorname{clip}\big(1, 1-\varepsilon, 1 + \varepsilon\big) A^{\text{GRPO}}_i \Big) \nabla \log \pitheta(\bm{y}^i|\bm{x})\Big] \\
& = \underset{\bm{x}^i \sim \pitheta}{\mathbb{E}}\Big[\frac{1}{N} \sum_{i = 1}^{N} A_i^{\text{GRPO}} \nabla \log \pitheta(\bm{y}^i|\bm{x})\Big] \\
& = \nabla\underset{\bm{x}^i \sim \pitheta}{\mathbb{E}}\Big[\frac{1}{N} \sum_{i = 1}^{N} A_i^{\text{GRPO}} \log \pitheta(\bm{y}^i|\bm{x})\Big]
\end{align*}
This is exactly policy gradient objectives with advantage $A_i = A_i^{\text{GRPO}}$

\paragraph{Online DPO} For Online DPO \cite{guo2024direct}, and its variants, Online SiLC \cite{zhao2023slic}, we generate the preference pairs from self-generated rollouts using the following protocol. After obtaining a group of responses $\bm{y}^1, \dots, \bm{y}^N$, we re-order the responses so that the reward values $r_1, \dots, r_N$ are monotonically decreasing as the response index grows. This is to say, we ensure $r_1 \geq r_2 \geq \dots \geq r_N$. Then, we create a preference pair data by matching $\bm{y}^1$ with $\bm{y}^N$, $\bm{y}^2$ with $\bm{y}^{N-1}$, etc, where we consider $\bm{y}^1, \bm{y}^2$ as the positive data and $\bm{y}^{N-1}, \bm{y}^{N}$ as the negative data. Since the online DPO-type objectives are computed for a pair of positive and negative data points, we illustrate their derivation using the notation $\bm{y}^+, \bm{y}^{-}$. Note that the DPO objective is given as
\begin{align*}
J_{\text{DPO}}(\theta)
= \mathbb{E}_{\bm{y}^+, \bm{y}^- \sim \pitheta(\cdot|\bm{x})} \Big[\log \sigma\!\big(z_\theta(\bm{x},\bm{y}^+,\bm{y}^-)\big)\Big],
\end{align*}
where $\sigma$ is the sigmoid function that is defined as $\sigma(x) = \frac{1}{1 + e^{-x}}$, and it satsifies $\sigma(-x) = 1 - \sigma(x)$, and $z_{\theta}$ is define as 
\begin{align*}
z_\theta(\bm{x},\bm{y}^+,\bm{y}^-) = \beta\Big(
\log \pi_\theta(\bm{y}^+ \mid \bm{x}) - \log \pi_\theta(\bm{y}^- \mid \bm{x})
-\log \pi_{\mathrm{ref}}(\bm{y}^+ \mid \bm{x}) + \log \pi_{\mathrm{ref}}(\bm{y}^- \mid \bm{x})
\Big).
\end{align*}
Therefore, computing the gradient of the DPO objective gives,
\begin{align*}
\nabla J_{\text{DPO}}(\theta)
&= \mathbb{E}_{\bm{y}^+, \bm{y}^- \sim \pitheta(\cdot|\bm{x})}\Big[ \nabla \log \sigma(z_\theta)\Big]
\\
&= \mathbb{E}_{\bm{y}^+, \bm{y}^- \sim \pitheta(\cdot|\bm{x})}\Big[ \frac{\partial}{\partial z_\theta}\log \sigma(z_\theta)\;\nabla  z_\theta\Big]
\\
&= \mathbb{E}_{\bm{y}^+, \bm{y}^- \sim \pitheta(\cdot|\bm{x})}\Big[ \big(1-\sigma(z_\theta)\big)\;\nabla  z_\theta\Big]
\\
&= \mathbb{E}_{\bm{y}^+, \bm{y}^- \sim \pitheta(\cdot|\bm{x})}\Big[ \big(\sigma(-z_\theta)\big)\;\beta\,
\nabla\Big(
\log \pi_\theta(\bm{y}^+ \mid \bm{x}) - \log \pi_\theta(\bm{y}^- \mid \bm{x})
\Big)\Big]
\\
&= \mathbb{E}_{\bm{y}^+, \bm{y}^- \sim \pitheta(\cdot|\bm{x})}\Big[ \beta \cdot \sigma(-z_\theta)\,
\Big(
\nabla \log \pi_\theta(\bm{y}^+ \mid \bm{x})
-\nabla \log \pi_\theta(\bm{y}^- \mid \bm{x})
\Big)\Big] \\
&= \mathbb{E}_{\bm{y}^+, \bm{y}^- \sim \pitheta(\cdot|\bm{x})} \Big[ \beta\cdot \sigma(-z_{\theta}) \nabla \log \pitheta(\bm{y}^{+} | \bm{x}) + (-\beta \cdot \sigma(-z_{\theta})) \nabla \log \pitheta(\bm{y}^{-} | \bm{x})\Big] \\
&= \nabla ~ \mathbb{E}_{\bm{y}^+, \bm{y}^- \sim \pitheta(\cdot|\bm{x})} \Big[ \beta\cdot \sigma(-z_{\theta}) \log \pitheta(\bm{y}^{+} | \bm{x}) + (-\beta \cdot \sigma(-z_{\theta}))  \log \pitheta(\bm{y}^{-} | \bm{x})\Big] 
\end{align*}
This is the same as weighted policy gradient objectives with advantage $\beta \cdot \sigma(-z_{\theta})$ assigned to $\bm{y}^+$ and $-\beta \cdot \sigma(-z_{\theta})$ assigned to $\bm{y}^-$.

\paragraph{Online DPO-smooth} We can also create a smoothed version of Online DPO to alleviate label noise arising from inaccurate preference pairs, which stem from the inherent flaws of the reward models. Let $\varepsilon$ be a label smooth/noise parameter, indicating that with probability $\varepsilon$, the obtained preference is wrong. Then, after taking this into consideration, the correct gradient for Online DPO-smooth should be
\begin{align*}
    \nabla J_{\text{DPO-smooth}} = \nabla ~ \mathbb{E}_{\bm{y}^+, \bm{y}^- \sim \pitheta(\cdot|\bm{x})} \Big[ \beta \cdot \Big(\big((1-\varepsilon) \cdot \sigma(-z_{\theta}) - \varepsilon \sigma(z_{\theta}) \big)\log \pitheta(\bm{y}^{+} | \bm{x}) - \big((1-\varepsilon) \cdot \sigma(-z_{\theta}) - \varepsilon \sigma(z_{\theta}) \big) \log \pitheta(\bm{y}^{-} | \bm{x}) \Big) \Big] 
\end{align*}
This is the same as weighted policy gradient objectives with advantage $(1-\varepsilon) \cdot \sigma(-z_{\theta}) - \varepsilon \sigma(z_{\theta})$ assigned to $\bm{y}^{+}$ and $-\big((1-\varepsilon) \cdot \sigma(-z_{\theta}) - \varepsilon \sigma(z_{\theta}) \big)$ assigned to $\bm{y}^-$.

\paragraph{Online SLiC} Similarly, we can compute derive for SLiC \cite{zhao2023slic}, 

\begin{align*}
J_{\mathrm{SLiC}}(\theta)
:= \mathbb{E}_{\bm{y}^+, \bm{y}^- \sim \pitheta(\cdot|\bm{x})} 
\Big[
-\max\big(0,\; \tau - \log \pi_\theta(\bm{y}^+ \mid \bm{x})
+ \log \pi_\theta(\bm{y}^- \mid \bm{x})\big)
\Big].
\end{align*}
where $\tau$ is a pre-defined threshold value. Define the margin violation indicator
\begin{align*}
\mathbb{I}_{\mathrm{viol}} = \mathbf{1}\!\left[
\tau - \log \pi_\theta(\bm{y}^+ \mid \bm{x})
+ \log \pi_\theta(\bm{y}^- \mid \bm{x}) > 0
\right].
\end{align*}
Therefore, the gradient of the objective
\begin{align*}
\nabla  J_{\mathrm{SLiC}}(\theta)
&= \mathbb{E}_{\bm{y}^+, \bm{y}^- \sim \pitheta(\cdot|\bm{x})} \Big[
\mathbb{I}_{\mathrm{viol}}\,
\nabla
\Big(
\log \pi_\theta(\bm{y}^+ \mid \bm{x})
- \log \pi_\theta(\bm{y}^- \mid \bm{x})
\Big)
\Big]
\\
&= \mathbb{E}_{\bm{y}^+, \bm{y}^- \sim \pitheta(\cdot|\bm{x})} \Big[
\mathbb{I}_{\mathrm{viol}}\,
\Big(
\nabla \log \pi_\theta(\bm{y}^+ \mid \bm{x})
-\nabla \log \pi_\theta(\bm{y}^- \mid \bm{x})
\Big)
\Big].
\end{align*}
This is the same as the gradient of the weighted policy gradient objective with advantage $\mathbb{I}_{\mathrm{viol}}$ assigned to $\bm{y}^{+}$ and advantage $-\mathbb{I}_{\mathrm{viol}}$ assigned to $\bm{y}^-$
\paragraph{SFT and Best-of-N self-distillation } It's straightforward to see that, for SFT, the gradient is the same as policy gradient with constant advantage value $1$ across all samples as the objectives are equivalent. For Best-of-N self-distillation \cite{sessa2024bond}, the objective is given as 
\begin{align*}
\mathcal{L}_{\text{distill}}(\theta) = \operatorname{KL}(\pi_{\text{BoN}} || \pitheta) = \mathbb{E}_{\bm{y}^i \sim \pi_{\text{BoN}}(\cdot | \bm{x})}\Big[ \log \dfrac{\pi_{\text{BoN}}(\bm{y}^i | \bm{x})}{\pitheta(\bm{y}^i |\bm{x})}\Big]
\end{align*}
Therefore, we can write its gradient as,
\begin{align*}
    \nabla \mathcal{L}_{\text{distill}}(\theta) & = \nabla \mathbb{E}_{\bm{y}^i \sim \pi_{\text{BoN}}(\cdot | \bm{x})}\Big[ \log \dfrac{\pi_{\text{BoN}}(\bm{y}^i | \bm{x})}{\pitheta(\bm{y}^i |\bm{x})}\Big] = \mathbb{E}_{\bm{y}^i \sim \pi_{\text{BoN}}(\cdot | \bm{x})} \Big[  \nabla \log \pitheta(\bm{y}^i | \bm{x})\Big]
\end{align*}
Since $\bm{y}^i \sim \pi_{\text{BoN}}(\cdot | \bm{x})$ represents that $\bm{y}^i$ is the one with highest reward $r_i$ among $r_1, \dots, r_N$, the objective can be simplified to weighted policy gradient objective with advantage $1$ assigned to the best sequence with highest reward $\bm{y}^i$ and other wise $0$. This is equivalent to performing SFT only on the self-generated best sequence.

\subsection{Answer Forcing}
\label{sec:appendix_answer_forcing}
In this section, we provide a detailed account of the proposed answer-forcing algorithm. This technique is applicable to tasks with verifiable rewards, where the reward is computed by checking the generated answer against a ground truth, such as the 0-1 correctness reward for math problem and IoU reward for object grounding. 

Given a group size of $N$, the naive implementation of answer-forcing described in Section \ref{sec:method_answer_forcing} would first generate $N$ samples, evaluate the rewards, and then decide whether to generate an additional sample via injection. This is highly inefficient. Instead, we always generate $N+1$ samples in parallel for each group, with 1 sample containing a ground-truth answer. However, depending on the rewards of the first $N$ samples, we optionally discard the extra sample from the loss computation when the remaining $N$ samples already include outputs with a significantly effective training signal (e.g., high rewards as measured by accuracy or IoU).

Concretely, during the online sampling process, we are given a prompt $x$, ground truth answer $z^*$ and a reward function $R$. When the desired group size is $N$, we always generate $N+1$ samples with one extra answer-forced sample. Specifically, we initialize timestamps $t_1=t_2=...t_N=1$ and  initialize $\bm{y}^1=\bm{y}^2..\bm{y}^N$ using a fully masked sequence ``\texttt{M M ...M}". We initialize $\bm{y}^{N+1}_{t_{N+1}}=\text{``}\texttt{M\,M\,}\dots\texttt{\,M\;} 
\texttt{<answer>}~z^\star~\texttt{</answer>}\text{"}$ with the answer section pre-filled. $t_{N+1}$ is set to the value $t'$ according to the mask ratio. For example, if the answer has 3 tokens and max generation length is 12, the timestep $t'$ will be $0.75$, since $\frac{9}{12}$ of the tokens are masked. 

All of these sequences have equivalent length, which is set to 512 for math reasoning, 128 for object grounding and 256 for image editing based on the distribution of reasoning lengths in our SFT data. Ideally, we do not want the generated sequences to have exactly 512, 256, or 128 tokens to allow for some flexibility. In standard sampling, the model generates special [PAD] tokens at the end of the sentence if the reasoning length is less than the maximum sequence length. For the answer-forced sample, we adopted the Fill-in-the-Middle (FIM) design of LaViDa \cite{li2025lavida} and inserted random-length ``\texttt{[S]...[S]}" sequences in the SFT data right before ``\texttt{<answer>...</answer>}", where [S] is a special infilling token. This design allows flexible-length text infilling as the model can generate ``\texttt{[S]...[S]}" if the reasoning length is less than the lengths of mask segment ``\texttt{[M]...[M]}" in $\bm{y}^{N+1}$.

After obtaining $\bm{y}^1_{t_1}...\bm{y}^{N+1}_{t_{N+1}}$, we perform diffusion sampling using the policy model $\pi_\theta$ to obtain $\bm{y}^i \sim \pi_\theta(\bm{y}^i_0 \mid \bm{x}, \bm{y}^i_{t_i})$ for $i=1,2...N,N+1$. Notably, the model forward computation of these samples can be performed in parallel. We then evaluate the rewards on these $N+1$ samples to obtain $r_1...r_{N+1}$. Finally, we check if the maximum rewards $r_\text{max}$ among non-answer-forced samples exceeds a threshold $\tau$. In our setup, $\tau$ is set to 0.5 for both 0-1 math rewards and IoU rewards. We do not include any auxiliary rewards such as format rewards in this step. If  $r_\text{max}$ exceeds $\tau$, we consider the original sample to have high-quality outputs and discard the answer-forced sample. Otherwise, we randomly replace one samples in $\bm{y}^1...\bm{y}^{N}$ with $\bm{y}^{N+1}$. Since $\bm{y}^1...\bm{y}^{N}$ are just i.i.d samples, we always discard $\bm{y}^1$ in our implementation. The answer-forcing algorithm is formally documented in Algorithm \ref{alg:answer_forcing}.

\begin{algorithm}[t]
\caption{Answer Forcing with dLLMs}
\label{alg:answer_forcing}
\begin{algorithmic}[1]
\Require Prompt $\bm{x}$, policy $\pi_\theta$, group size $N$, ground-truth answer $z^\star$,
reward function $R$, threshold $\tau$, injection ration $\beta$
\Ensure Training group $\mathcal{G}$

\State Initialize forced sample:

\For{$i=1$ to $N$}
    \State $\bm{y}^{i}_1\gets ``\texttt{M\,M\,}\dots\texttt{\,M\;} "$
    \State $t_i \gets 1$
\EndFor

\State $\bm{y}^{N+1}_{t'}\gets ``\texttt{M\,M\,}\dots\texttt{\,M\;} 
\texttt{<answer>}~z^\star~\texttt{</answer>}"$

\State $t_{N+1} \gets t' $

\For{$i=1$ to $N+1$ \textbf{in parallel}}
    \State $\bm{y}^i \sim \pi_\theta(\bm{y}^i_0 \mid \bm{x}, \bm{y}^i_{t_i})$
\EndFor

\For{$i=1$ to $N$}
    \State $r^i \gets R(\bm{x}, \bm{y}^i,z^*)$
\EndFor
\State $r_{\max} \gets \max_{i \in [N]} r^i$

\If{$r_{\max} < \tau$ and $Rand() < \beta$}

\State $\mathcal{G} \gets \{(\bm{x}, \bm{y}^i)\}_{i=2}^{N+1}$
    
\Else
    \State $\mathcal{G} \gets \{(\bm{x}, \bm{y}^i)\}_{i=1}^N$
\EndIf

\State \Return $\mathcal{G}_{\text{loss}}$
\end{algorithmic}
\end{algorithm}

\subsection{Tree Search}
\label{sec:appendix_tree_search}

In this section, we provide detailed descriptions of our tree-search algorithm. This technique is applicable to tasks without ground-truth answers but with a real-valued reward function. It is not applicable to 0-1 rewards, since we cannot meaningfully identify a best sample when all rewards are zero.

Given a prompt $x$ and reward function $R$, and a base group size $N$, the tree search process is controlled by the number of tree expansions $k$ and max diffusion steps $T$, and restart timestep index $s_1...s_k\in\{0,1..T\}$. In particular, the restart timestep index determines which point in the saved trajectories should serve as the branching point. $s_1$ is always 0 since we always need to go through the full $T$ diffusion steps for the first $N$ samples $\bm{y}^1...\bm{y}^N$. The indices $s_1...s_k\in\{0,1..T\}$ directly correspond to diffusion timesteps $t_1...t_N\in[0,1]$ through the relation $t_i=1-\frac{s_i}{T}$.

% \begin{equation}
    
%     \label{eq:timestep_relation}
% \end{equation}. 

In particular, $t_1=1$ always holds, indicating fully masked sequences.

After the $i$th batch is generated, we will have $Ni$ samples. We find the index $m$ corresponding to the sample with the highest reward among all previously generated samples $\bm{y}^1.\bm{y}^{Ni}$. In the $(i+1)$th batch, we generate the $(Ni+1)$th sample to the $(Ni+N)$ th sample using $\bm{y}^m_{t_{i+1}}$ as the starting point.  These samples will go through $T-s_{i+1}$ diffusion steps. This process is repeated until all $Nk$ samples are generated.

In a distributed training setup, samples in each batch are generated on multiple GPUs in parallel. Hence, we need to gather the generated trajectories and evaluated rewards. To reduce the cost of maintaining and synchronizing multiple trajectories in memory, we use a more compact representation, leveraging the fact that once a token is unmasked at a diffusion step, it will not be modified in subsequent steps. 

Concretely, we store the final generated result $\bm{y}^j_0$ and an array $v^j\sim\{1,..T\}^L$ which keeps track of when each token is unmasked, where $L$ is the sequence length. $v^j[i]=p$ indicates the i-$th$ token is unmasked at p-$th$ diffusion step where $1\le p \le T$. To recover  $\bm{y}^j_{t'}$ for arbitrary $t'\in[0,1]$, we can obtain the corresponding diffusion step index $s'$ through the relationship $t_i=1-\frac{s_i}{T}$ and recover $\bm{y}^j_{t'}$ through $\bm{y}^j_{t'}[i]=\bm{y}^j_0[i]$ if $v^j[i]\le s'$ and $\bm{y}^j_{t'}[i]=M$ otherwise. This reduces the overhead from $\mathcal{O}(NLT)$ to $\mathcal{O}(NL)$ at each batch. The tree search algorithm is formally described in Algorithm \ref{alg:tree_search}.

\begin{algorithm}[t]
\caption{Tree Search with dLLMs}
\label{alg:tree_search}
\begin{algorithmic}[1]
\Require Prompt $\bm{x}$, policy $\pi_\theta$, reward function $R$, base group size $N$, number of tree expansions $k$, diffusion steps $T$, restart timestep index $s_1...s_k\in\{0,1,...T\}$ with $s_1=0$
\Ensure Generated group $\mathcal{G}$ of size $Nk$

\State Initialize empty group $\mathcal{G} \gets \emptyset$

\For{$i=1$ to $K$}
    \State $t_{i} \gets 1-\frac{s_i}{T}$ \Comment{We always have $t_1=1-\frac{s_1}{T}=1$}
\EndFor

\For{$j=1$ to $N$}

    \State $\bm{y}^{j}_{t_1} \gets ``\texttt{M\,M\,}\dots\texttt{\,M}$" \Comment{Fully masked sequence}
\EndFor

\For{$i=1$ to $k$}
% \State $o \gets N(i-1)$ \Comment{Start index, this batch is the $(o+1)^{th},...,(o+N)^{th}$ samples.}
    \For{$j=Ni-N+1$ to $Ni$ \textbf{in parallel}}
        
        \State $\bm{y}^{j}_0 \sim \pi_\theta(\bm{y}_0 \mid \bm{x}, \bm{y}^{j}_{t_i})$
        \State $r^{j} \gets R(\bm{x}, \bm{y}^{j}_0)$
        \State Record unmasking order $v^{j} \in \{1,\dots,T\}^L$
    \EndFor
    \State Gather $\{r^{j},v^{j},\bm{y}^{j}\}_{j=Ni-N+1,Ni-N+2,...Ni}$ across processes
        \State $\mathcal{G} \gets \mathcal{G} \cup \{(\bm{x}, \bm{y}^j_0)\}_{j=Ni-N+1,Ni-N+2,...Ni}$ \Comment{$\mathcal{G}$ has $Ni$ samples after this point.}
    \State $m \gets \arg\max_{j \in \{1,2..{Ni}\}} r^j$
\State $\bm{y}^m_{t_{i+1}}[\ell] \gets 
\bm{y}^m_0[\ell]\;\textbf{if}\; v^m[\ell] \le s_{i+1},\ \textbf{otherwise}\ \texttt{M},
\quad \forall \ell = 1,\dots,L$ \Comment{Recover restart state}

    \For{$j=Ni+1$ to $Ni+N$}
        \State $\bm{y}^{j}_{t_{i+1}} \gets \bm{y}^m_{t_{i+1}}$ \Comment{Prepare next batch for restarting from partial state}
    \EndFor
\EndFor

\State \Return $\mathcal{G}$ \Comment{$\mathcal{G}$ has $Nk$ samples.}
\end{algorithmic}
\end{algorithm}

\begin{figure}
    \centering
    \includegraphics[width=0.8\linewidth]{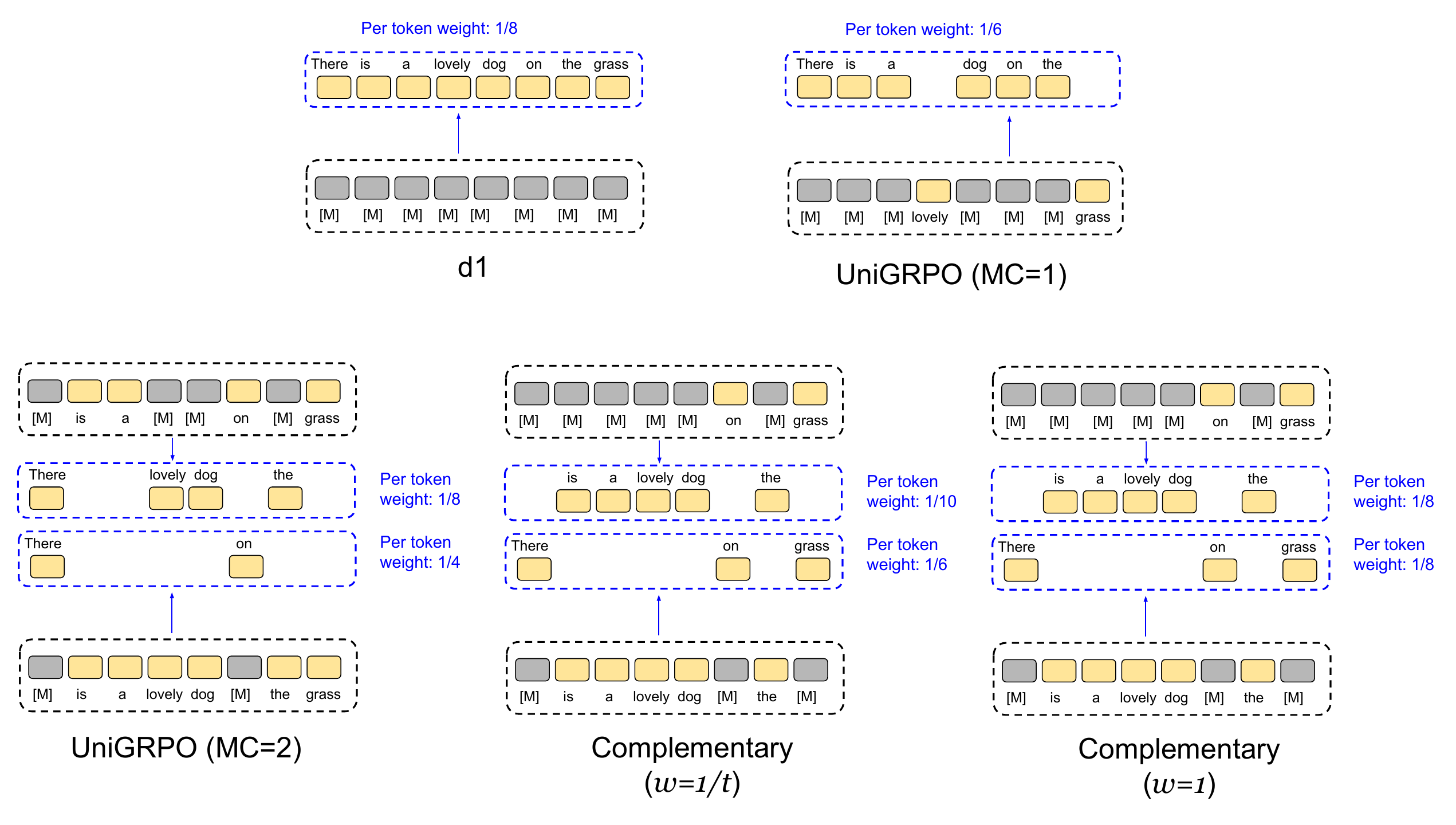}
    \caption{\textbf{Different Design Choices of Likelihood Estimators.} a) d1 uses MC=1, $w(t)=\frac{1}{t}=1$ and always sample $t=1$, equivalent to fully masked sequences. Hence, it can compute per-token likelihood at all positions.  b) UniGPRO uses MC=1 and $w(t)=\frac{1}{t}$. It randomly masks a subset of tokens and computes the likelihood with only masked tokens. c) We can improve UniGRPO by increasing the MC to 2 for a broader token coverage. However, it leads to imbalanced gradients with $w(t)=\frac{1}{t}$. d) In vanilla complementary masking developed for pretraining, two MC samples are coupled to ensure 100\% token coverage. However, the token imbalance issue persists because it uses $w(t)=\frac{1}{t}$. e) In our design, we set $w(t)=1$, ensuring all-token coverage while balancing the importance of each token in a uniform fashion.  } 
    \label{fig:likelihood}
\end{figure}
\subsection{\textbf{Likelihood Estimator}}
\label{sec:likelihood}
In this section, we discuss more technical details of likelihood estimation in dLLMs. Recall that we estimate the data log-likelihood using the ELBO, as follows. 
\begin{align*}
\log \pitheta(\bm{y} |\bm{x}) = \mathbb{E}_{t, \bm{y}_t}\Big[w(t)\sum_{k \in \{k| \bm{y}_t[k]=\M\}} \log \pitheta(\bm{y}[k] |\bm{y}_t, \bm{x})\Big]
\end{align*}
We will list several ELBO estimation methods from the literature for comparison with ours. A detailed visualization is presented in Figure \ref{fig:likelihood}.
\paragraph{d1 \cite{zhao2025d1}} d1 estimates the likelihood by constantly sampling $t = 1$ and $\bm{y}_1 = [\M, \dots, \M]$, uses only one sample (MC = 1), with weight $w(t) = \frac{1}{t}$. This gives the estimation expression as,
\begin{align*}
\log \pitheta^{\text{d1}}(\bm{y} |\bm{x}) = \sum_{k=1}^L \log \pitheta(\bm{y}[k] |\bm{y}_1, \bm{x})
\end{align*}
\paragraph{UniGRPO \cite{yang2025mmada}} UniGRPO instead sample mask ratio $t$ from $[0,1]$ uniformly and a corresponding binary mask $\mathsf{M}$ associated with $\bm{y}_t$ based on such mask ratio (MC = 1). It adopts the weight $w(t) = \frac{1}{t}$. This means that it computes the data loglikelihood with,
\begin{align*}
\log \pitheta^{\text{unigrpo}}(\bm{y} |\bm{x}) = \frac{1}{t}\sum_{k \in \{k| \bm{y}_t[k]=\M\}} -\log \pitheta(\bm{y}[k] |\bm{y}_t, \bm{x})
\end{align*}
We could improve the design by increasing the Monte Carlo sample size to 2 (MC=2) by sampling another masked sample with a different mask rate, chosen also uniformly at random. 
\begin{align*}
\log \pitheta^{\text{unigrpo}, 2}(\bm{y} |\bm{x}) = \frac{1}{2}\Big[ \frac{1}{t_1}\sum_{k \in \{k| \bm{y}_{t_1}[k]=\M\}} \log \pitheta(\bm{y}[k] |\bm{y}_{t_1}, \bm{x}) + \frac{1}{t_2}\sum_{k \in \{k| \bm{y}_{t_2}[k]=\M\}} \log \pitheta(\bm{y}[k] |\bm{y}_{t_2}, \bm{x})\Big]
\end{align*}

% \paragraph{DMPO \cite{zhu2025enhancing}} DMPO uses $4$ samples (MC=4) for estimating the log probability by sampling $t_1, 1-t_1, t_2, 1-t_2$ as the time. It also ensures the use of a complementary mask, that is $\mathsf{M}^1 = 1 - \mathsf{M}^2$, $\mathsf{M}^3 = 1 - \mathsf{M}^4$. With $\mathsf{M}^1, \mathsf{M}^2, \mathsf{M}^3, \mathsf{M}^4$ associated samples $\bm{y}_{t_1}$, $\bm{y}_{1- t_1}$, $\bm{y}_{t_2}$, $\bm{y}_{1-t_2}$. It still uses weighting $w(t) = \frac{1}{t}$ , giving the following expression,
% \begin{align*}
% \log \pitheta^{\text{DMPO}}(\bm{y} |\bm{x}) & = \frac{1}{4}\Big[ \frac{1}{t_1}\sum_{k: \bm{y}_{t_1}[k]=\M} \log \pitheta(\bm{y}[k] |\bm{y}_{t_1}, \bm{x}) + \frac{1}{1- t_1}\sum_{k: \bm{y}_{1-t_1}[k]=\M} \log \pitheta(\bm{y}[k] |\bm{y}_{1-t_1}, \bm{x}) \\
% & + \frac{1}{t_2}\sum_{k: \bm{y}_{t_2}[k]=\M} \log \pitheta(\bm{y}[k] |\bm{y}_{t_2}, \bm{x}) +  \frac{1}{1 - t_2}\sum_{k: \bm{y}_{1-t_2}[k]=\M} \log \pitheta(\bm{y}[k] |\bm{y}_{1-t_2}, \bm{x})\Big]
% \end{align*}

\paragraph{Our recipe } We adopt a simple, effective approach for estimating the log probability using $2$ samples with complementary masking (MC=2). We sample $\bm{y}_{t_1}\sim q(y_{t_1}|y_0)$ using the discrete forward diffusion process, and set

\begin{equation}
\bm{y}_{t_2}[i] =
\begin{cases}
\bm{y}[i], & \text{if } \bm{y}_{t_1}[i] = M, \\
M,         & \text{if } \bm{y}_{t_1}[i] \neq M.
\end{cases}
\end{equation}

Finally, we compute the likelihood via
 
\begin{align*}
\log \pitheta^{\text{ours}}(\bm{y} |\bm{x}) = \frac{1}{2}\Big[ \sum_{k \in \{k| \bm{y}_{t_1}[k]=\M\}} \log \pitheta(\bm{y}[k] |\bm{y}_{t_1}, \bm{x}) + \sum_{k \in \{k| \bm{y}_{t_2}[k]=\M\}} \log \pitheta(\bm{y}[k] |\bm{y}_{t_2}, \bm{x})\Big]
\end{align*}

This approach differs from vanilla complementary masking used by a few other works \cite{li2025lavida, zhu2025enhancing,bie2025llada2} in that we adopt $w(t) = 1$  as opposed to $w(t) = \frac{1}{t}$. While this seems like a small modification, it has a profound impact on model performance, as confirmed by the ablation study in \ref{sec:timestep_weighting_appendix}.

We visualize these likelihood estimators in Figure \ref{fig:likelihood}.

% 

% In our design, we use two samples with timestep $t_1\sim \text{Uniform}([0,1])$ and $t_2=1-t$. We sample $\bm{y}_{t_1}\sim q(y_{t_1}|y_0)$ using the discrete forward diffusion process, and set $\bm{y}_{t_2}[i]=\bm{y}[i]$ if $\bm{y}_{t_1}[i]=M$ and  $\bm{y}_{t_2}[i]=M$  if $\bm{y}_{t_1}[i]\ne M$. This design, known as complementary masking, was first proposed in LaViDa\cite{li2025lavida} for pretraining. For example, if the sequence is $
% \bm{y}$ is ``\texttt{there is a dog}" and $
% \bm{y}_{t_1}$ is ``\texttt{[M] is [M] dog}", $\bm{y}_{t_2}$ will be ``\texttt{there [M] a [M]}".  We adopt a $w(t)=1$ instead of  $w(t)=\frac{1}{t}$ from LaViDa, giving the following estimator

\section{Additional Experiment Details and Results }
\label{sec:appendix_setup}
\subsection{Setup}

\textbf{Training Dataset}

Our training consists of two stages. Stage 1 consists of only SFT objective. Stage 2 incorporates both online sampling and offline datasets in a unified fashion.  

Stage 1 datasets can be generally categorized into two categories. The first category is a subset of LaViDa-O's pretraining data, which includes:

\begin{itemize}
    \item \textit{A: Text-to-Image Pairs.} We source data from LAION-2B \citep{schuhmann2022laion}, COYO-700M \citep{kakaobrain2022coyo-700m}, BLIP3o-60k \citep{chen2025blip3}, ShareGPT4o-Image \citep{chen2025sharegpt}. Each dataset is heavily filtered to remove NSFW prompts, low CLIP scores \citep{radford2021learning}, low aesthetic scores \citep{laion-aesthetics}, and low-resolution images following LaViDa-O's pipeline. We include all data from BLIP3o-60k and ShareGPT4o-Image, and select highest-quality samples from LAION and COYO based on CLIP scores and aesthetic scores.  This part consists of 20M images.
    
    \item \textit{B: Image-level Understanding Data.} We include MAmmoth-VL \citep{guo2024mammoth}, and VisualWebInstruct \citep{visualwebinstruct}. 
    \item \textit{C: Region-level Understanding Data.} We include GranD \citep{hanoona2023GLaMM} and RefCOCO \citep{kazemzadeh2014referitgame}.
    \item \textit{D: Image Editing Data.} We include ShareGPT4o-Image \citep{chen2025sharegpt}, GPT-Edit-1.5M \citep{wang2025gpt}, and the image editing subset of UniWorld-V1 \citep{hu2022unified}.
 \end{itemize}

 The second category includes newly incorporated reasoning data not used in LaViDa-O's original training recipe. It includes

\begin{itemize}
    \item \textit{E: Visual Understanding with Reasoning.} We include Vision-R1-Cold\cite{huang2025vision}, which consists of visual math problems and VQA problems.
    \item \textit{F: Pure Language-based Reasoning.} We include DeepScalar \cite{deepscaler2025}. 
    \item \textit{G: Image Editing with Reasoning.} We include GoT \cite{fang2025got} data. Additionally, we run the data pipeline of GoT on GPT-Edit-1.5M \citep{wang2025gpt} to further expand the data.
    \item \textit{H: Reason-intensive Grounding.} We include ReasonSeg \citep{lai2024lisa} and Lisa-CoT \citep{yang2023lisa++}.
 \end{itemize}

Given the substantial imbalance between the two categories, we design a dataloader to prioritize the acquisition of reasoning capabilities. The dataset of the two categories is sampled with a 3:7 ratio.

Stage 2 training includes all datasets from Stage 1. It also incorporates additional RL datasets, including 

\begin{itemize}
    \item \textit{I: Visual Understanding with Reasoning.} We include Vision-R1-Cold and Vision-R1-RL \cite{huang2025vision}, which consists of visual math problems and VQA problems, and ViRL-39k \cite{wang2025vl}
    \item \textit{J: Language Based Reasoning.} We include DeepScalar \cite{deepscaler2025}, GSM8K \cite{cobbe2021gsm8k}, MATH \cite{lightman2023lets}. 
    \item \textit{K: Image Editing with Reasoning.} We include input images and prompts from EditScore-RL \cite{luo2025editscore}.
    \item \textit{L: Reason-intensive Grounding.} We include ReasonSeg \citep{lai2024lisa}, Lisa-CoT \citep{yang2023lisa++},  RefCOCO \citep{kazemzadeh2014referitgame}.
 \end{itemize}

 While some datasets such as DeepScalar \cite{deepscaler2025} appear both in the SFT data and RL data, they are incorporated in different manners. In RL data,  the reasoning traces provided in these datasets are not used. Only the prompts and corresponding answers are used.

\textbf{Reward Functions}
For math problems and question answering, we use 0-1 correctness reward. For object grounding, we use IoU reward. For image editing, we use a VLM-based reward model, Editscore\cite{luo2025editscore}. Since Editscore requires considerable memory, it is hosted on a separate server to the main training processes. In addition to these key rewards, we also incorporate auxiliary rewards such as tag-formatting rewards and a repetition penalty, following existing work \cite{gao2024designing,zhao2025d1}. We note that the formatting reward in existing works like d1 is not robust as it merely counts the ``$<$think$>$" and ``$<$answer$>$" tokens. We replace it with a more robust formatting reward based on regular expressions.

\textbf{Training Hyperparameter}

 We report the training hyperparameters, including the learning rate, number of training steps, optimizer setup, and image resolution for understanding and generation tasks in Table \ref{tab:training-stages} for reference. The training is conducted of 64 GPUs. The main experiment is conducted on H100s. Some ablation studies use A100s. The total training time takes 5 days for Stage 1 and 3 days for Stage 2. We observe a considerable bottleneck in reward evaluation. In particular, with a global batch size of 256 images, the edit score evaluation takes 70-140 seconds. We will investigate how to further optimize this infrastructure bottleneck in future work by improving the network infrastructure and scaling reward servers.  
 
\begin{table}[h]
\centering
\caption{\textbf{Training configurations of \ours.} We report the relevant hyperparameters for training, including the learning rate, number of training steps, optimizer setup, image resolution for understanding and generation tasks. }
\label{tab:training-stages}
\begin{tabular}{lHHcc}
\toprule

 & \textbf{Stage 1} & \textbf{Stage 2} & \textbf{Stage 1} & \textbf{Stage 2} \\
\midrule
Learning Rate & $5 \times 10^{-6}$ & $1 \times 10^{-4}$ & $5 \times 10^{-6}$ & $5 \times 10^{-7}$ \\
Steps & 80k & 400k & 100k  & 5k \\
$\beta_1$ & 0.99 &0.99  & 0.99 & 0.99 \\
$\beta_2$ & 0.999 &0.999  & 0.999  & 0.999  \\
optimizer & AdamW & AdamW & AdamW  & AdamW \\
\midrule
Dataset Used & B,C & A & A,B,C,D,E,F,G,H & A,B,C,D,E,F,G,H,I,J,K,L \\
Loaded Parameters & 8B & 6.4B & 10.4B  & 10.4B \\
Trainable Parameters & 8B & 2.4B & 10.4B & 10.4B  \\
Und. resolution & 384 $\times \{(1,3),(2,2)\}$ & 384 $\times \{(1,3),(2,2)\}$ & 384 $\times \{(1,3),(2,2)\}$  & 384 $\times \{(1,3),(2,2)\}$ \\
Gen. resolution & - & 256 $\rightarrow$ 512 $\rightarrow$ 1024 & 1024 & 1024 
\\
Loss & - & 256 $\rightarrow$ 512 $\rightarrow$ 1024 & SFT & Unifed \\
\midrule
% Semantic Encoder & Trainable & Not Loaded & Trainable \\
% VQ Encoder & Not Loaded & Loaded & Loaded \\
% Gen. Branch & Not Loaded & Trainable & Trainable \\
% Und. Branch & Trainable & Partially Loaded & Trainable \\
% \bottomrule
\end{tabular}%

\end{table}

\textbf{Evaluation Protocol.} We use a max sequence length of 128 for object grounding, 256 for image editing and 512 for all other tasks to match the training setup. All other sampling parameters remain unchanged following LaViDa-O's inference protocol.

\subsection{Additional Ablation Studies}

In this section, we report additional ablation study results to verify the design of \ours.

\subsection{Timestep Weighting}
\label{sec:timestep_weighting_appendix}

In Table \ref{tab:ablation_liklihood}, we report results of varying the timestep weighting discussed in Section \ref{sec:likelihood}. Specifically, we compare results of complementary masking with $w=\frac{1}{t}$ and $w=1$, results show that $w=1$ works better. Specifically, we observe that $w=\frac{1}{t}$ significantly degrades performance on image-editing tasks. This may be attributed to the large number of tokens in visual generation tasks, which exacerbates the imbalance in per-token loss. In particular, image editing tasks produce 256 text tokens and 4096 image tokens per sample, significantly more than understanding-only tasks, which generate only 512 tokens.
 
\subsection{Alternative Losses in the Unified Framework}
\begin{table}[]
    \centering
        \caption{\textbf{Ablation Studies of Per-token Likelihood Weighting }}
\begin{tabular}{H cHH c c}
\toprule
\textbf{N-MC} & \textbf{Weighting} & \textbf{Source} & \textbf{Math500} & \textbf{Lisa-Grounding} & \textbf{ImgEdit} \\
\midrule
% 1 & i.i.d           & UniGRPO     & TODO & 61.9 & 3.82 \\
% 1 & Full    & d1          & TODO & 59.2 & 3.77 \\
2 &LaViDa-O + SFT & UniGRPO     & TODO & 40.3  & 3.81 \\
2 & $w(t)=\frac{1}{t}$ & UniGRPO     & TODO & 64.3 & 3.82 \\
2 & $w(t)=1$   & LaViDa-R1   & 38   & \textbf{65.0} & \textbf{3.90} \\
\bottomrule
\end{tabular}

    \label{tab:ablation_liklihood_weight}
\end{table}
\begin{figure}[t]
    \centering
    \includegraphics[width=1.0\linewidth]{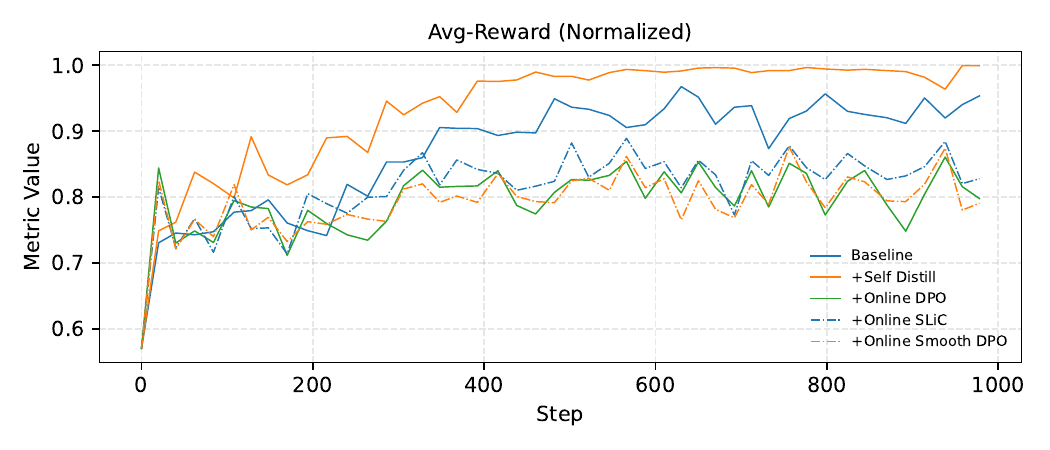}
    \caption{\textbf{Ablation Studies of Combining Different Losses in the Unified Framework}}
    \label{fig:dpo_loss}
\end{figure}
\subsection{Qualitative Results}
\label{sec:appendix_qualitative_results}
In Figure \ref{fig:dpo_loss}, we visualize the reward trajectories of combining RL loss with other post-training methods beyond best-of-N distillation discussed in Section \ref{sec:appendix_unified}. Specifically, we start with the GRPO+SFT baseline and explore incorporating additional losses, including online DPO, online DPO-smooth, and online SLiC loss. Notably, all these can be achieved by simply modifying the weight $A_i$ in Equation \ref{eq:awc_loss} without considerably changing the training pipeline. Results show that adding other losses is worse than the GRPO+SFT baseline, while incorporating Best-of-N distillation improves upon the baseline. Although most auxiliary losses did not improve performance, this experiment highlighted the flexibility of our proposed framework. We hope future work will further explore loss-weighting recipes beyond group advantages in GRPO.

\begin{figure}
    \centering
    \includegraphics[width=0.8\linewidth]{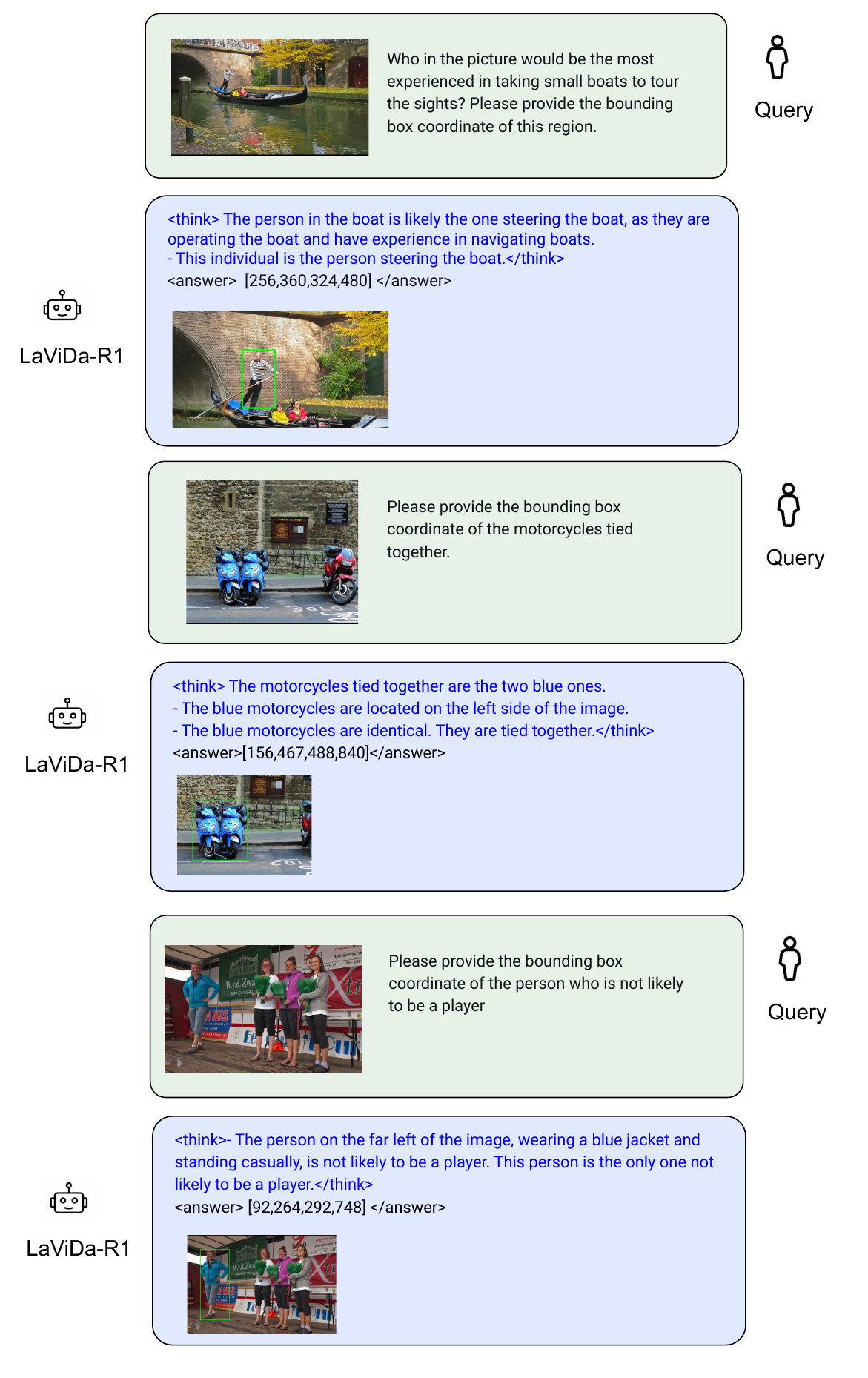}
    \caption{\textbf{Qualitative Results of Reason-Intensive Object Grounding.}}
    \label{fig:demo_1}
\end{figure}

\begin{figure}
    \centering
    \includegraphics[width=0.8\linewidth]{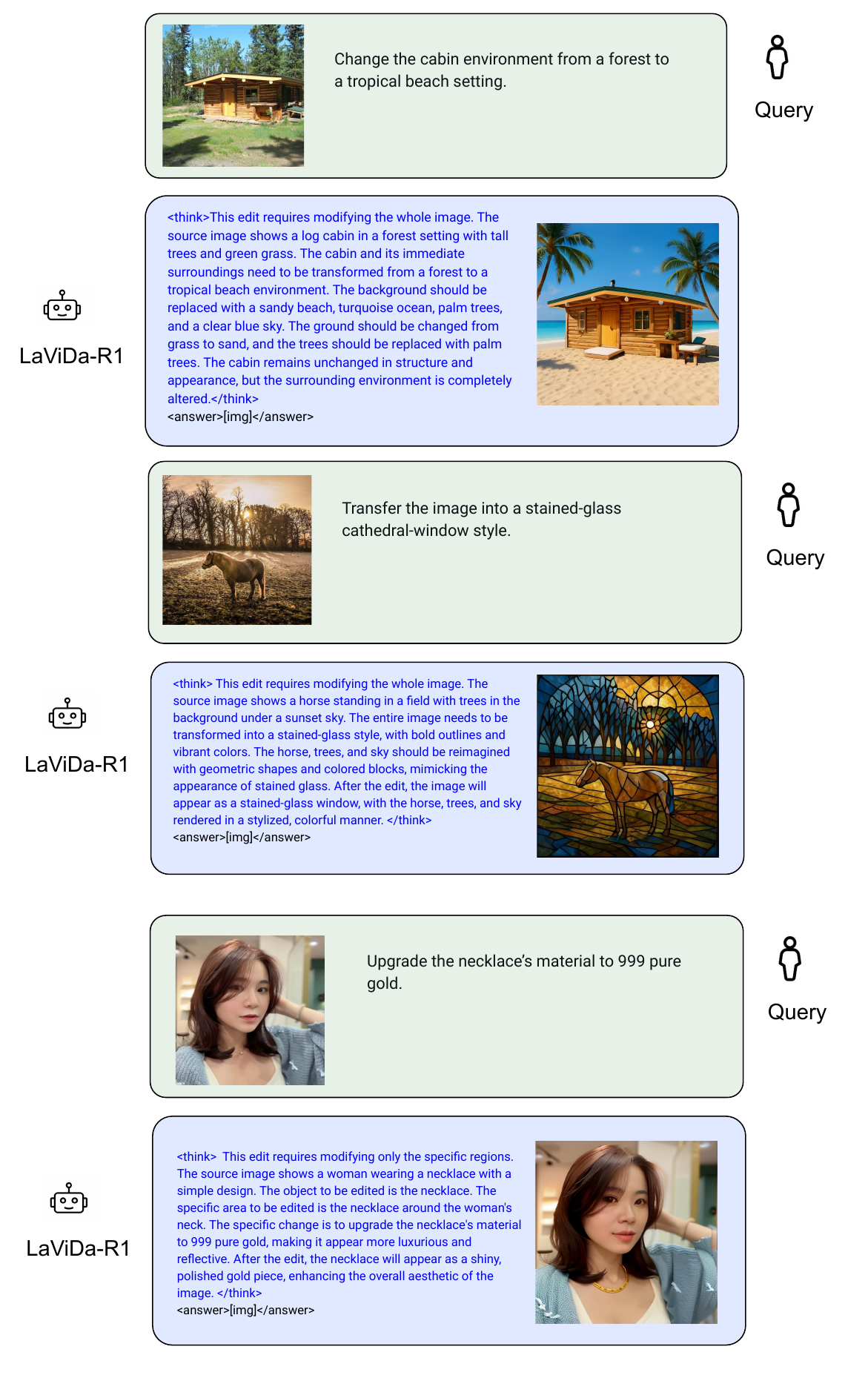}
    \caption{\textbf{Qualitative Results of Reason-Based Image Editing.}}
    \label{fig:demo_2}
\end{figure}

\begin{figure}
    \centering
    \includegraphics[width=0.8\linewidth]{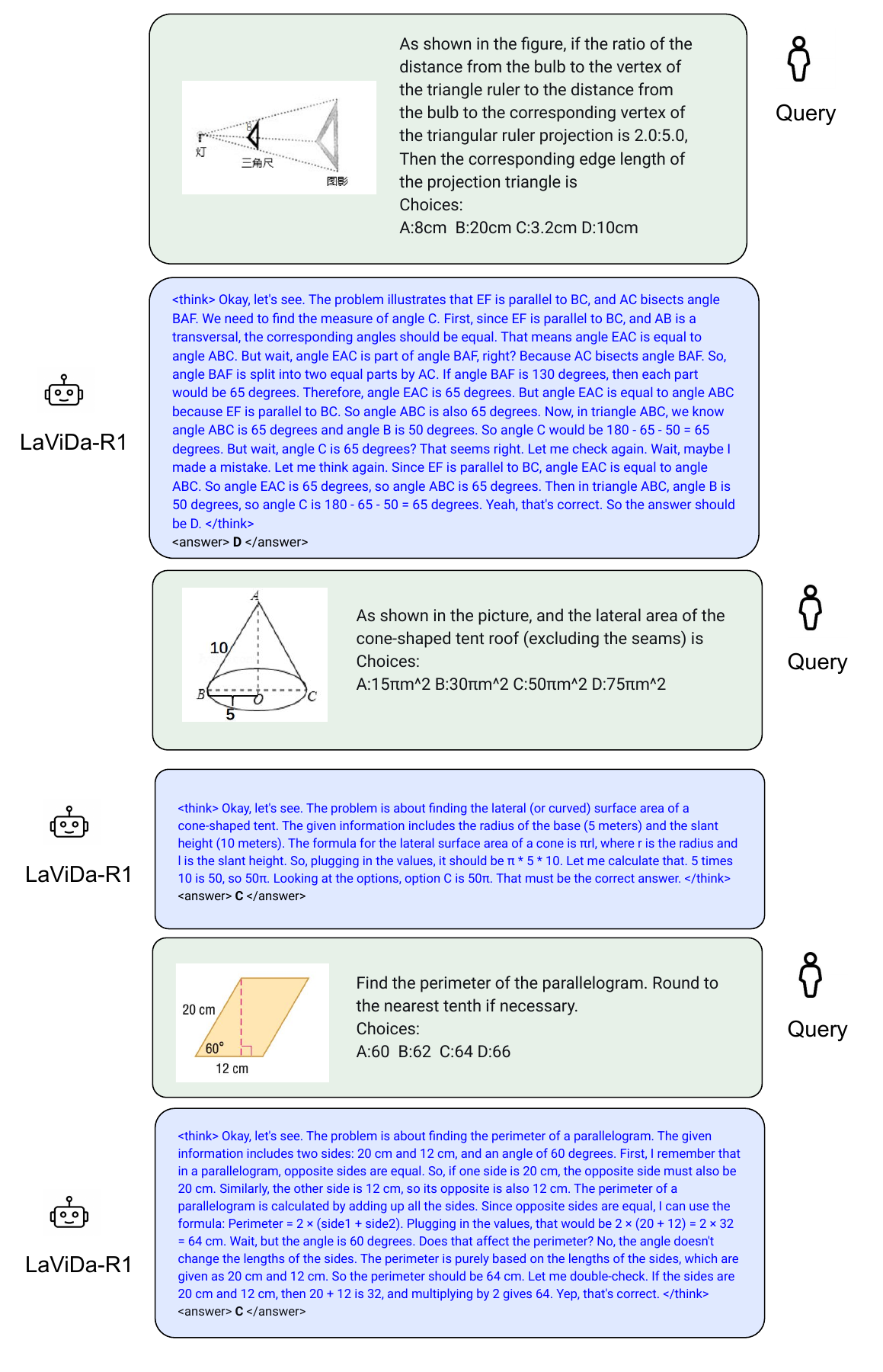}
    \caption{\textbf{Qualitative Results on Visual Math Problems.}}
    \label{fig:demo_3}
\end{figure}

To further demonstrate the effectiveness of \ours, we include qualitative results of model outputs on object grounding tasks in Figure \ref{fig:demo_1}, image editing tasks in Figure \ref{fig:demo_2} and math reasoning tasks in Figure \ref{fig:demo_3}. These results demonstrated the strong reasoning capabilities of \ours~on diverse multimodal tasks.

\section{Limitation}

Despite the strong results of \ours, it has several key limitations. First, while \ours~improves upon the base model LaViDa-O. There still exists a considerable gap between the reasoning performance of multimodal dLLMs and state-of-the-art AR MLLMs such as Qwen3-VL \cite{Qwen3-VL}. We hope that future work on scaling and improving pretraining can close this gap for base models. Second, while AR LLMs can leverage many efficient inference frameworks such as vLLM \cite{kwon2023efficient}, these low-level optimization frameworks have not yet been fully adapted to support dLLMs. This results in a throughput bottleneck because our training process uses eager Python execution during online sampling. We hope future acceleration frameworks tailored to dLLMs will help address this issue. Lastly, while \ours~already included diverse number of tasks and significantly expanded the scope of dLLM RL literature, there are more tasks to explore for unified multimodal dLLMs, such as multi-turn website code generation with visual feedback, interleaved text-image reasoning, poster design, etc. In this work, we focused on single-round reasoning with visual inputs. We will explore more tasks in future work.

\begin{figure}
    \centering
    \includegraphics[width=0.8\linewidth]{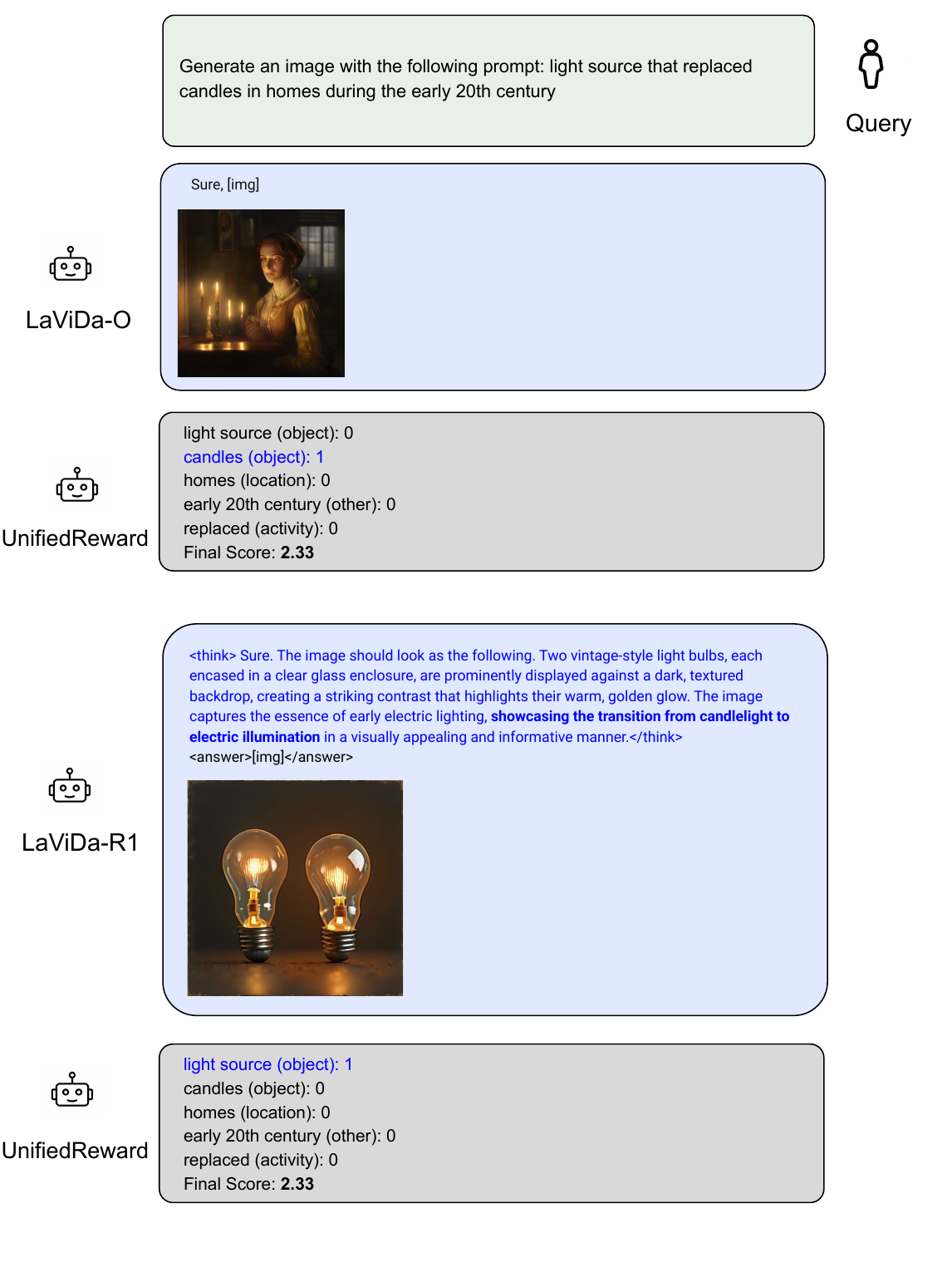}
    \caption{\textbf{Qualitative Reuslts on Text-to-Image Generation.} While \ours~demonstrated some zero-shot reasoning capabilities on text-to-image tasks, we find that existing VLM-based reward models fail to properly provide a reward signal for reasoning-based tasks. }
    \label{fig:demo_t2i}
\end{figure}

\textbf{Text-to-Image Generation}.We also briefly explored extending reasoning to text-to-image (T2I) generation. However, we find that existing reward models are fundamentally misaligned with reasoning-centric objectives. Despite this limitation, \ours\ demonstrates non-trivial zero-shot reasoning ability on T2I tasks. An illustrative example is shown in Figure~\ref{fig:demo_t2i}: given the prompt “The light source that replaced candles in homes during the early 20th century,” \ours\ correctly reasons about the historical context and generates light bulbs, whereas LaViDa-O produces images of candles.

Unfortunately, current reward models fail to reliably distinguish such reasoning-grounded generations. Early approaches such as PickScore~\cite{kirstain2023pick}, which are built upon CLIP-based models, lack explicit reasoning capabilities. We further investigated more recent VLM-based reward models, but even the state-of-the-art UnifiedReward-Qwen-7B~\cite{unifiedreward} is unable to correctly rank the samples in this example. Upon inspecting the model’s justifications, we find that it hallucinates spurious criteria, assessing alignment with surface-level concepts such as “candles” (object) and “replaced” (activity), thereby misinterpreting the compositional and historical reasoning required by the prompt.

Recent work~\cite{wei2025skywork} proposes using a frontier VLM (GPT-4.1) as an online reward model, which may alleviate some of these issues. However, this approach is prohibitively expensive and difficult to scale. We believe that effective and scalable reward modeling remains a key open challenge for reasoning-driven T2I generation. In light of these limitations, we leave reinforcement learning fine-tuning for T2I reasoning to future work.

% For example, in the WISE benchmark \cite{niu2025wise}, the prompts involve complex world knowledge such as ``\texttt{Einstein's favorite musical instrument}" or ``\texttt{The Pyramids of Giza at 8 PM Tokyo time.}". 
% However, existing reward models like HPSv3\cite{ma2025hpsv3} and PickScore \cite{kirstain2023pick} are not trained on such distribution of prompts and are unfit to evaluate prompt-image alignment for these cases. Most recent works have explored 

\section{Additional Discussions for Related Works}
\label{sec:appendix_related}

\textbf{Reinforcement Learning with dLLMs.} d1 \cite{zhao2025d1} first explored adapting the GPRO algorithm through token log probability ratios estimated through one-sample ELBO, with many works exploring other improved designs of estimation \cite{gong2025diffucoder, yang2025mmada, wang2025d2}. In another line of work, wd1 \cite{tang2025wd1} and DMPO \cite{zhu2025enhancing} addressed the dLLM RL tasks from a policy distribution-matching perspective, leading to new ways of computing advantage that differ from the aforementioned GRPO-style algorithms. TraceRL \cite{wang2025revolutionizing} considers an actor-critic style method, and ESPO \cite{ou2025principled} adapts GSPO \cite{zheng2025group} to dLLMs.

\textbf{Unified SFT and RL.} There have been multiple works in the LLM community that explored effective strategies to combine SFT and RL into a single stage \citep{chen2025retaining, lv2025towards, liu2025uft, fu2025srft, yan2025learning, ma2025learning}. These works have introduced approaches that dynamically balance off-policy SFT loss and on-policy RL loss, attempting to outperform single-stage RL or single-stage SFT for LLM post-training. Unlike these approaches, our loss formulation for unified SFT and RL is simple and was introduced to ensure the stability of the RL stage, rather than to make single-stage RL or single-stage SFT obsolete. Moreover, to the best of our knowledge, our work is the first to apply a unified SFT-RL strategy to unified multimodal models, whereas prior work has largely focused on single-text modalities using LLMs. 

\textbf{Connection between Answer Forcing and IGPO.} IGPO \cite{zhao2025inpainting} injects partially masked ground-truth reasoning traces as a hint in the online sampling process of dLLMs. While this design and our proposed answer forcing algorithm both make use of inpainting capabilities of dLLMs, they are distinct from each other in that IGPO requires high-quality ground-truth reasoning traces, whereas Answer Forcing only requires access to the final answer and does not require any pre-existing reasoning traces. In general, high-quality reasoning traces are rare and difficult to obtain for tasks such as object grounding, making answer forcing a more generalizable approach.

% \section{Broader Impact}

% \section{}

% \section{You \emph{can} have an appendix here.}

% You can have as much text here as you want. The main body must be at most $8$
% pages long. For the final version, one more page can be added. If you want, you
% can use an appendix like this one.

% The $\mathtt{\backslash onecolumn}$ command above can be kept in place if you
% prefer a one-column appendix, or can be removed if you prefer a two-column
% appendix.  Apart from this possible change, the style (font size, spacing,
% margins, page numbering, etc.) should be kept the same as the main body.
% %%%%%%%%%%%%%%%%%%%%%%%%%%%%%%%%%%%%%%%%%%%%%%%%%%%%%%%%%%%%%%%%%%%%%%%%%%%%%%%
% %%%%%%%%%%%%%%%%%%%%%%%%%%%%%%%%%%%%%%%%%%%%%%%%%%%%%%%%%%%%%%%%%%%%%%%%%%%%%%%

\end{document}